\documentclass[11pt,a4paper]{article}
\usepackage[hyperref]{emnlp2020}
\usepackage{times}
\usepackage{latexsym}
\usepackage{footnote}

\usepackage{microtype}

\aclfinalcopy 
\def\aclpaperid{1388} 

\usepackage{amsmath,amsfonts,bm}

\def\eqref#1{equation~\ref{#1}}

\def\1{\bm{1}}

\def\rvx{{\mathbf{x}}}

\DeclareMathAlphabet{\mathsfit}{\encodingdefault}{\sfdefault}{m}{sl}
\SetMathAlphabet{\mathsfit}{bold}{\encodingdefault}{\sfdefault}{bx}{n}

\DeclareMathOperator*{\argmin}{arg\,min}

\usepackage{multirow}
\usepackage{algorithm}
\usepackage[noend]{algpseudocode}
\usepackage[export]{adjustbox}
\usepackage{booktabs}
\usepackage{subfig}

\usepackage{stfloats}

\newcommand{\xiangdone}[1]{}
\newcommand{\junyidone}[1]{}
\newcommand{\xisendone}[1]{}

\newcommand{\nop}[1]{}

\newcommand{\para}[1]{\smallskip\noindent\textbf{#1}}

\usepackage{titlesec} 
\titlespacing*{\section}{1pt}{8pt}{2pt}
\titlespacing*{\subsection}{1pt}{8pt}{2pt}
\titlespacing*{\subsubsection}{1pt}{1pt}{1pt}

\newcommand{\vcoll}{VisCOLL}

\title{Visually Grounded Continual Learning of Compositional Phrases}

\author{Xisen Jin\quad Junyi Du\quad Arka Sadhu\quad Ram Nevatia\quad Xiang Ren \\
Department of Computer Science, University of Southern California\\
\texttt{\{xisenjin, junyidu, asadhu, nevatia, xiangren\}@usc.edu} \\
}

\begin{document}
\maketitle

\begin{abstract}
Humans acquire language continually with much more limited access to data samples at a time, as compared to contemporary NLP systems.
To study this human-like language acquisition ability,
we present \vcoll, a visually grounded language learning task, which simulates the continual acquisition of compositional phrases from streaming visual scenes.
In the task, models are trained on a paired image-caption stream which has shifting object distribution; while being constantly evaluated by a visually-grounded masked language prediction task on held-out test sets.
\vcoll{} compounds the challenges of \textit{continual learning} (\textit{i.e.}, learning from continuously shifting data distribution) and \textit{compositional generalization} (\textit{i.e.}, generalizing to novel compositions).
To facilitate research on \vcoll{}, we construct two datasets, COCO-shift and Flickr-shift, 
and benchmark them using different continual learning methods.
Results reveal that SoTA continual learning approaches provide little to no improvements on \vcoll{}, since storing examples of all possible compositions is infeasible.
We conduct further ablations and analysis to guide future work~\footnote{Code and data: \url{https://github.com/INK-USC/VisCOLL}}.


\end{abstract}

\section{Introduction}
Modern NLP systems, including ones that build on pre-trained language models~\cite{Devlin2019BERTPO,radford2019language},
excel on a wide variety of tasks.
These systems rely on offline (batch) training and have drawn recent criticism due to their inability to adapt to new contexts~\cite{linzen2020generalization}.
In contrast, humans acquire language from evolving environments, require a small memory footprint~\cite{mcclelland1995there}, and can generalize their knowledge to newer tasks~\cite{sprouse2013comparison}.
It has been suggested that humans ground perceptual experience to semantically interpret symbols \cite{Bisk2020ExperienceGL,harnad1990symbol,vigliocco2014language}.

\begin{figure}[tb]

\hspace{-0.2cm}
    \centering
    \includegraphics[width=\linewidth]{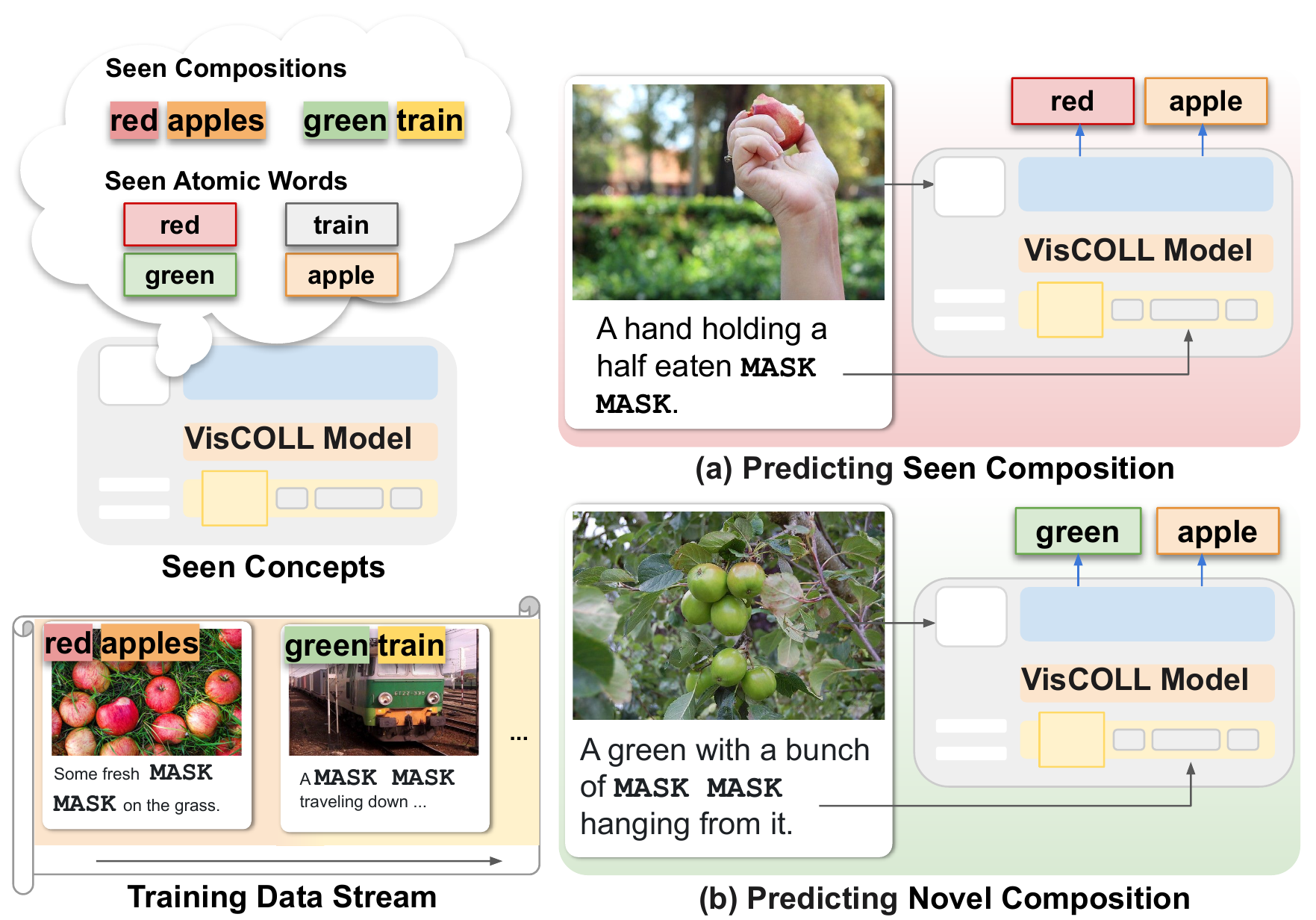}

    \caption{\textbf{Illustration of the proposed \vcoll task.}
    The end-task is Masked-Token Prediction: given an image, a model predicts the masked tokens of an associated caption in an online continual learning setup (cf. (a) in the figure).
    Additionally, we test composition generalization by evaluating on novel compositions (b) which are not encountered at train time.
    }
    \label{fig:task}
    \vspace{-0.2cm}
\end{figure}

To model the challenge, we propose \vcoll, a \textbf{Vis}ually-grounded \textbf{Co}ntinua\textbf{L} \textbf{L}earning setup, to acquire compositional phrases from streaming visual-linguistic data.
Models receive a stream of paired image-caption data which has a shifting object distribution.
As the end task, we employ masked token prediction of captions given the associated image, as illustrated in Fig.~\ref{fig:task}(a). This evaluates a model's learned knowledge on composing phrases with the given context.


\vcoll{} captures two \textit{inter-related} challenges.
First, unlike previous continual learning works on image classification~\cite{kirkpatrick2017overcoming, zenke2017continual}, \vcoll{} requires predicting, for example, a noun with a verb or an adjective---which results in a significantly large search space.
As a result of this increased search space, memory based continual methods~\cite{Robins1995CatastrophicFR, Aljundi2019OnlineCL} cannot expect to store prototypes of each visited compositions.
Second, the increased search space makes it infeasible to view all possible combinations of atomic words at train time.
Therefore, to succeed on \vcoll, models should
generalize to novel compositions at test time
(also called \textit{composition generalization}) ~\cite{lake2017generalization, keysers2020measuring}.

In this work, we extensively study the challenges associated with \vcoll.
To facilitate the research, we 
construct a continuously shifting data distribution to closely resemble real-word data-stream and contribute 
COCO-shift and Flickr-shift.
We benchmark these datasets using multi-modal language modeling architectures \cite{Tan2019LXMERTLC,Su2020VL-BERT:} which achieve state-of-art performance on multiple vision-language tasks.
In particular, we don't use any pre-training, instead train randomly initialized models on streaming data using continual learning algorithms \cite{Robins1995CatastrophicFR, rolnick2019experience, Aljundi2019OnlineCL} and evaluate their resistance to forgetting and compositional generalization. 
We quantify the performance and forgetfulness of trained models and evaluate on a novel test split to measure compositional generalization, as shown in Fig.~\ref{fig:task}(b).




Our proposed \vcoll{} benchmark reveals that the gains
observed in image classification tasks from state-of-art continual learning algorithms fail to transfer to \vcoll{} even with increased memory.
On the other hand, composition generalization remains challenging even for offline-training.

To summarize, our contributions are: 
(i) we propose \vcoll{}, a task aimed at continual learning of compositional semantics from visually grounded text inputs
(ii) we contribute two datasets COCO-shift and Flickr-shift to study \vcoll{} and benchmark them with multi-modal language models 
(iii) we show that existing continual learning algorithms fail at  learning compositional phrases and provide potential future research direction.

\section{Related Works}

\noindent
\textbf{Continual Learning.} 
A major challenge in continual learning is to alleviate catastrophic forgetting~\cite{Robins1995CatastrophicFR}. Several recent works ~\citep{Greco2019PsycholinguisticsMC, wang2019sentence, dAutume2019EpisodicMI} study the challenge in the context of NLP.
Existing continual learning algorithms can be broadly classified into memory-based approaches~\cite{LopezPaz2017GradientEM, Aljundi2019GradientBS}, pseudo-replay based approaches~\cite{shin2017continual}, regularization based approaches~\cite{kirkpatrick2017overcoming, zenke2017continual, v.2018variational} and architecture based approaches ~\cite{rusu2016progressive}. 
However, these works are mainly designed for image classification tasks where the training data has ``clear" task boundaries--\textit{i.e.,} training stream are partitioned into disjoint subsequences.
In contrast, task boundaries in \vcoll{} are unknown and ``smooth" (i.e., with gradual transitions between tasks)--a setting that is closer to real-world situations.
Thus, \vcoll{} rules out many continual learning algorithms which require explicit 
task identity and boundary~\cite{kirkpatrick2017overcoming, rusu2016progressive}.

\para{Modeling Language Compositionality.}
Capturing compositionality in language has been a long challenge~\cite{fodor1988connectionism} for neural networks.
Recent works explore the problem with
compositional generalization on synthetic instruction following~\cite{lake2017generalization}, text-based games~\cite{yuan2019interactive}, visual question answering~\cite{bahdanau2019closure}, and visually grounded masked word prediction~\cite{suris2019learning}.
In particular, ~\citet{Li2020Compositional} study a closely related task of continual learning of sequence prediction for synthetic instruction following. However, their techniques for separating semantics and syntax is restricted to text-only case. 
\citet{nguyen2019contcap} investigate continual learning of image captioning, but make strong assumptions on the structure of the data stream, and do not evaluate compositionality.

Different from these, our work focuses on learning compositional language (\textit{e.g.}, phrases) in a continual learning setup. We create realistic training streams to simulate shifting data distribution, with systematic evaluation of compositionality learned in the models. 
We aim at improving model's ability of acquiring language from real-world streaming data with a low-memory footprint.


\begin{figure*}
    \hspace{-0.7cm}
    \includegraphics[width=1.04\textwidth]{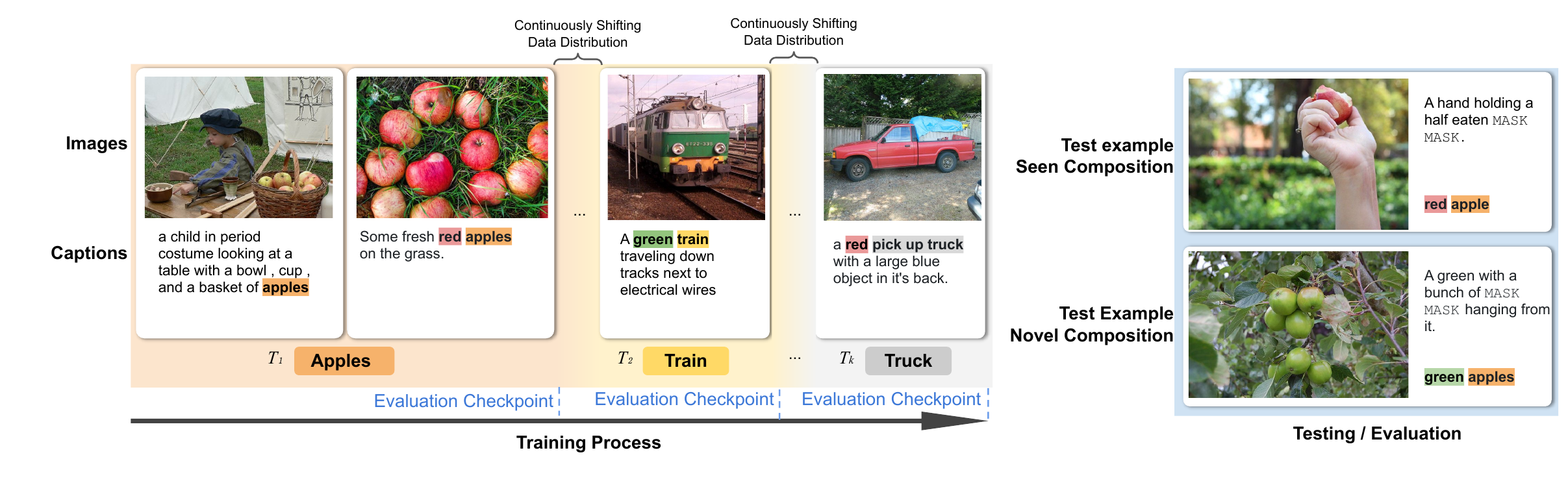}
    \vspace{-0.6cm}
    \caption{
    \textbf{An overview of training and testing process in VisCOLL.}
    At \textbf{train} time, the model receives a stream of masked captions (highlighted in text) with their associated image. 
    We use the noun appearing in the masked token as the ``task'' subsequently used to create a continuously shifting data distribution.
    We further evaluate the model's performance every fixed time interval to quantify ``forgetting''.
    At \textbf{test} time, the models receives a seen composition or a novel composition of seen words.
    }
    \label{fig:dataset}
\end{figure*}

\begin{figure}[t]
\vspace{-0.0cm}
    \hspace{-0.3cm}
    \includegraphics[width=0.52\textwidth]{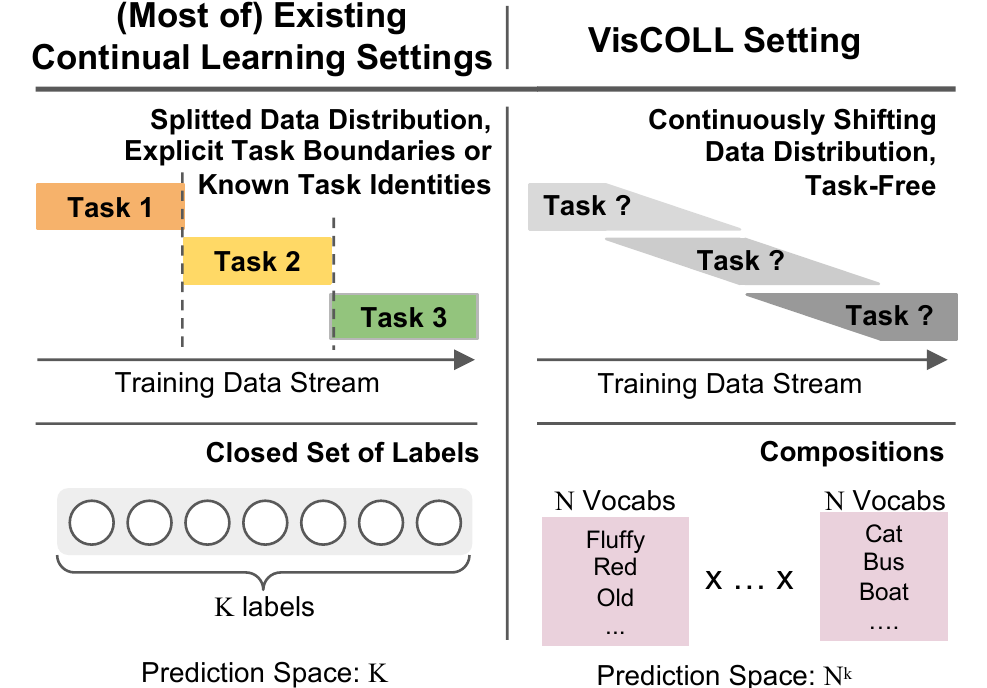}
    \vspace{-0.6cm}
    \caption{
    \textbf{Comparison of traditional continual learning setup for image classification with the setup of \vcoll.}
    In most of existing continual learning settings, task identities are obtainable and can be used to define task boundaries.
    In contrast, models in \vcoll{} is agnostic to task identities (task-free), and fed/trained by a gradually shifting data distribution. 
    }
    \label{fig:compare}
    \vspace{-0.0cm}
\end{figure}

\section{The VisCOLL Task}
\label{ss:task_def}
\noindent
\textbf{Overview.}
There are two design considerations for \vcoll{}:
\textit{compositionality} and \textit{continual learning}.
To test compositionality, we choose visually grounded masked language modeling where the model needs to compose atomic words to describe complex and novel visual scenes.
To simulate a realistic continual learning setup, we construct a dynamic environment where the training data comes in as a non-stationary data stream without clear ``task'' boundaries.
Since the goal is to simulate language acquisition from scractch, \vcoll{} models shouldn't initialize weights from a pre-trained network.
We introduce details of our task setup in the rest of the section. 

\para{Prediction End-Task.}
We employ visually grounded masked language modeling as the prediction task in VisCOLL.
An input instance to this task consists of an image $\rvx_{img}$, its object bounding boxes $\rvx_{bbox}$ (without object labels), and the caption text $\rvx_{text}$, where a contiguous text span in $\rvx_{text}$ is masked with \texttt{MASK} tokens which the model needs to fill.
The masked text span $\rvx_{label}$ always includes a noun and optionally includes verbs or adjectives. 
We define each noun, verb, and adjective as an atom, and evaluate the model in both ``seen'' and ``novel'' composition setting of nouns and verbs/adjectives. 
For instance, we may test whether the model successfully predicts ``\textit{red apples}'' (adjective+noun) when the model has seen examples that involve ``\textit{red}'' and ``\textit{apples}'' separately (see Figure~\ref{fig:dataset} for an illustration).

\para{Training Data Stream.} 
Unlike traditional offline training setups where a model  visits the training examples for multiple passes, we study an online continual learning setup, where the model visits a non-stationary stream of data.
We assume the data distribution changes gradually: for example, the model may see more ``\textit{apples}'' in the beginning, and see less of them later.
Unlike prior continual learning benchmarks, we do not assume strict task boundaries, \textit{i.e.}, sudden distribution changes.
We illustrate this distinction in Figure \ref{fig:compare}.

Formally, at each time step $t$, the model receives a small mini-batch of stream examples $\{\mathbf{X}_i\}_{i=0}^{B-1}$ where $X_i{=}(\rvx_{img}^{i}, \rvx_{bbox}^{i}, \rvx_{text}^i, \rvx_{label}^{i})$ whose distribution changes over time, \textit{i.e.,} $P(X_i, t_i) \neq P(X_i, t_j)$ where the time step $t_i \neq t_j$.
Note that our formulation rules out continual learning algorithms that make use of information about task boundaries.
Section ~\ref{sec:data} formally introduces the data stream construction process.


\para{Evaluation.}
In addition to the final performance, we also evaluate the model performance every fixed time interval and compute its forgetting as the performance loss over time.
For compositional generalization, a novel composition split is used.
Details will be covered in the following Sections~\ref{sec:data} and~\ref{sec:main_expt}.

\section{Dataset Construction}
\label{sec:data}


We construct our data streams using two popular vision-language datasets: COCO-captions \cite{Chen2015MicrosoftCC} and Flickr30k Entities \cite{Plummer2015Flickr30kEC} which provide multiple captions for each image in MSCOCO \cite{lin2014microsoft} and Flickr30k \cite{flickr30k} respectively.
We call the resulting datasets COCO-shift and Flickr-shift (see Table \ref{tab:stat} for dataset statistics).

Constructing a dataset for \vcoll{} involves two key steps:
(i) identify the phrase to be masked which involves a noun and associated verbs and adjectives
(ii) create a non-stationary data-stream.

\begin{table}[tb]
\centering
\scalebox{0.75}{\begin{tabular}{@{}lcc@{}}
\toprule
\textbf{Dataset}     & \textbf{COCO-shift}   & \textbf{Flickr-shift} \\ 
\midrule
Size of training set & 639,592 & 456,299 \\
Size of validation (dev) set & 27,918 & 15,406    \\
Size of regular test set    & 28,720  & 15,286  \\
Size of compositional test set & 4,426 & - \\
Mean of instance \# per task & 26,288 & 487 \\
Median of instance \# per task & 7,727 & 137 \\
Average masked span length & 2.497 & 3.380 \\
Number of tasks  & 80  & 1,000   \\ \bottomrule 
\end{tabular}
}
\vspace{-0.2cm}
\caption{\textbf{Statistics of our constructed datasets COCO-shift and Flickr-shift.}}
\label{tab:stat}
\end{table}

\para{Masking Tokens.} 
First, we append part-of-speech tags (POS) to each caption using Stanza \cite{qi2020stanza}.
For Flickr30k Entities, we use the annotated noun-phrases as mask tokens. 
For COCO-captions, we identify text spans with a regular expression chunker with the following regular expression.
\begin{align*}
   \texttt{CHUNK: <DT>?<JJ|VBG|VBN>*<NN|NNS>+} \\
 \texttt{<VB|VBD|VBG|VBN|VBP|VPZ>*}
\end{align*}
The resulting text span always includes a noun, and optionally include a determinator and an adjective and verb before or after the noun. 


To construct a data-stream, we define a ``task'' as the object being referred to in the textual input data.
For Flickr30k Entities, this is simply the lemmatized noun in the masked span.
For COCO-captions, we further map the lemmatized nouns to the 80 object categories defined in MSCOCO using a synonym table provided in~\cite{Lu2018Neural}.

\begin{figure}[t]
    \centering
    \includegraphics[width=\linewidth]{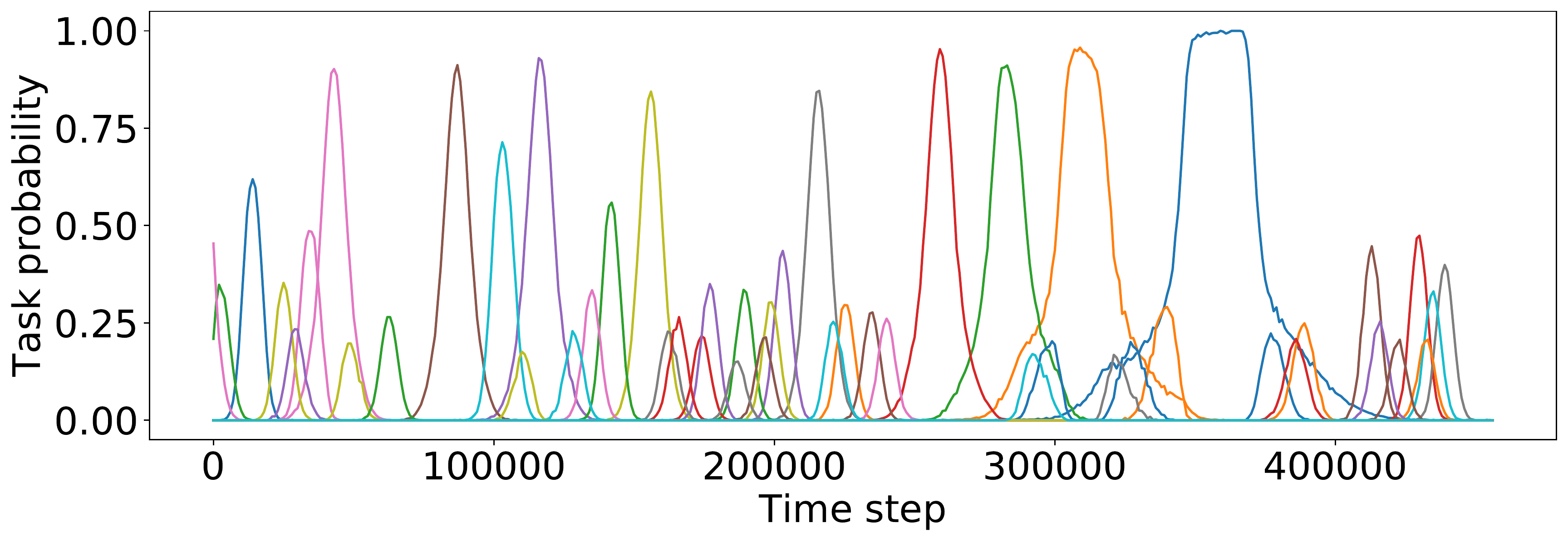}

    \caption{
    \textbf{Probability distributions of $50$ tasks (the noun in the masked tokens) in Flickr-shift data stream.}
    Each curve corresponds to a task. 
    $x$-axis shows the time step, and $y$-axis shows the probability of visiting the given task at a specific time step.
    }
    \label{fig:stream}

\end{figure}

\para{Non-Stationary Data-Stream.}
With a  set of ``tasks'' defined as 
$\mathcal{T}$, we let each task $T_i \in \mathcal{T}$ gradually introduced in the stream, then gradually fade out.
We generate a random permutation of all $K$ tasks $(T_{1}, T_{2}, T_{3}, ..., T_{K})$ as the order in which the centroid (mode) of each task distribution arrives in the stream. 
Each task proposes a task distribution for itself, which is a gaussian with $\mu_i = |D_i| / 2 +  \sum_{k<i} |D_k|$ and $\sigma_i = |D_i| / 2$, where $D_i$ is the set of training instances for task $i$. 
$\mu_i$ and $\sigma_i$ roughly determines the centroid and the spread of the distribution of each task. 
Finally, the algorithm greedily tries to put the proposed number of instances into each time interval to construct the stream. 
As a result, the constructed data stream has a gradually shifting task distribution without strict boundaries.
Figure~\ref{fig:stream} illustrates the task distribution in our constructed data streams. 

For COCO-shift, we separate out instances related to 5,000 images from the official validation set as our test set, and the rest as the test set. For Flickr-shift, we use the official train, validation and the test split. Note that the ``task'' is only used as an identifier of data distribution for constructing the dataset; the task identities are not revealed to models and the way we construct the data streams ensures there are no clear task boundaries.

\para{Test Split of Novel Compositions.} 
We measure compositional generalization by evaluating on a disjoint set of noun-adjective or noun-verb compositions.
We use the compositional test split of COCO dataset by~\citet{Nikolaus2019CompositionalGI} and remove images related to predefined 24 concept pairs (\textit{e.g.}, black cat, standing child) from the training, validation and the regular test set. The test split is referred to as the \textit{compositional} test set, and the rest is referred to as the \textit{regular} test set.



\begin{figure*}[tb]
    \centering
    \includegraphics[width=0.95\textwidth]{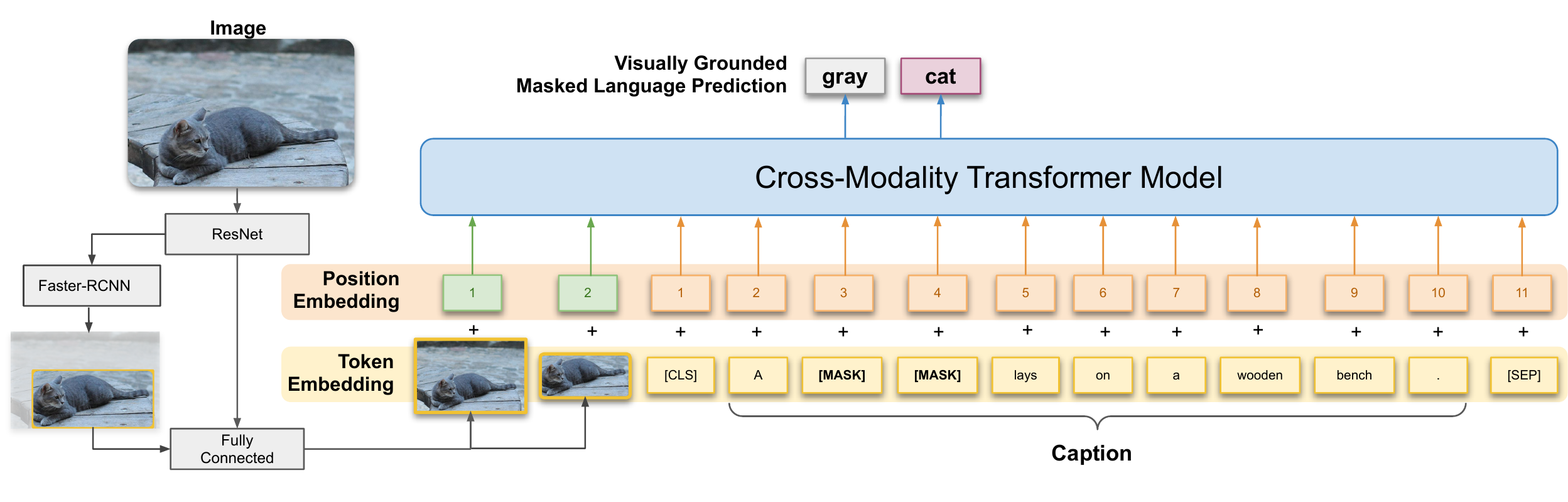}
    \caption{
    \textbf{Model architecture of the visual-language encoder used in \vcoll.}
    For the input image, we first extract image-level and object-level features using a FasterRCNN. 
    These features along with the masked caption are passed to a cross-modal transformer (in our case LXMERT and VLBERT) to predict the masked tokens.
    The model is \textit{randomly initialized without pre-trained weights}, and trained end-to-end with cross-entropy-loss.
    }
    \label{fig:arch}
\end{figure*}

\section{Methods}
\label{sec:exp_setup}
To benchmark on \vcoll{} and study the challenges it poses on model learning, we establish several continual learning baselines.
We use visual-language encoder models (Sec.~\ref{ss:base_model}) for masked token predictions. 
These models are trained from scratch (\textit{i.e.}, randomly initialized) with continual learning algorithms (Sec.~\ref{ss:cont_algos}) to dissuade catastrophic forgetting.


\subsection{Architectures of Visual Language Model}
\label{ss:base_model}
Recall that our end-task is masked token prediction where the input is an image and a caption with masked out tokens.
Since the task of masked token prediction is used as a pre-training method in almost all multi-modal masked language models, we choose two such model architectures, VLBERT~\cite{Su2020VL-BERT:} and LXMERT~\cite{Tan2019LXMERTLC}, as encoder models but expect similar conclusions with other models like ViLBERT \cite{Lu2019ViLBERTPT} or UNITER \cite{chen2019uniter}.
Since we seek to simulate the language acquisition process, 
we \textit{remove the pre-trained weights} from the models and randomly initialize the model weights for both the visual and language transformers.

For both base models, we first extract image and object features in ground truth bounding boxes with a Faster-RCNN~\cite{ren2015faster} model with Resnet-101 backbone ~\cite{He2015DeepRL} pretrained on Visual Genome ~\cite{krishna2017visual} dataset using only object-detection (without attributes).
The ground-truth object labels are not provided to the feature extractor model.
The extracted features are then passed to the base models along with the caption with masked text span replaced with \texttt{[MASK]} tokens.
Finally we compute the cross entropy loss with model predictions.
We illustrate our base model in Figure \ref{fig:arch}.


\subsection{Compared Methods}
\label{ss:cont_algos}
\para{Non-continual Learning Comparators.} The most common way of training a deep learning model is to perform gradient updates on a mini-batch of independently and identically distributed samples.
Since the model has access to the complete data (offline mode), the process is repeated multiple times (multiple epochs); we call this \textbf{offline} training.
To decouple the effect of training for multiple epochs, we report \textbf{offline (one pass)} where we restrict to a single pass over the data. Due to the absence of catastrohpic forgetting, the results are generally better than continual learning scenarios. Note that these two methods do not address the VisCOLL task and potentially indicate and upper-bound of performance.

\para{Continual Learning Methods.} In a continual learning setup like VisCOLL, the distribution of training examples is non-stationary, and due to limited memory only a part of the data can be stored.
In general, simply performing gradient updates after receiving a mini-batch (also called \textbf{Vanilla} Online Gradient Descent), leads to catastrophic forgetting~\cite{Robins1995CatastrophicFR}.


For \vcoll{}, we focus on memory-based continual learning algorithms.
These can be easily adapted to our setup as they don't require any task identifiers (not available in \vcoll).
We leave the exploration of regularization based approaches ~\cite{hsu2018re} to future work.

(1) \textit{Experience Replay} (\textbf{ER})
~\cite{Robins1995CatastrophicFR} randomly stores visited examples in a fix-sized memory called the replay memory, and these stored examples can later be randomly sampled for retraining. Similar techniques have been used in reinforcement learning~\citep{Schaul2016PrioritizedER, rolnick2019experience}. We apply reservoir sampling~\cite{vitter1985random} to decide examples to store and replace from the memory. The algorithm ensures each visited example has the same probability to be stored in the memory. 
(2) \textit{Average Gradient Episodic Memory} (\textbf{AGEM})
~\cite{chaudhry2018efficient}. Unlike ER, AGEM projects the gradients to a direction of non-increasing average loss computed on a random subset in the memory to alleviate forgetting.
(3) \textit{ER with Maximally Interfering Retrieval} (\textbf{ER-MIR})
~\cite{Aljundi2019OnlineCL} 
extends ER by selecting examples that are most likely to be forgotten in the next one update.

We further implement a method, \textbf{ER-MIR$_{max}$} as a variation of ER-MIR specific to our compositional prediction setting; which, instead of selecting the most likely forgotten phrase, selects the phrases containing the most likely forgotten words. It prevents the importance of an example get under-estimated when the example contains mostly easy words and a few forgettable words.





\begin{savenotes}
\begin{table*}[]
\centering

\resizebox{\textwidth}{!}{%
\begin{tabular}{@{}llcccccccccccc@{}}
\toprule
\toprule
& \textbf{Model}                         & \multicolumn{6}{c}{\textbf{VLBERT}}                                                                                                   & \multicolumn{6}{c}{\textbf{LXMERT}}                                                                                            \\ \cmidrule(r){1-2} \cmidrule(lr){3-8} \cmidrule(lr){9-14}
& \textbf{Metrics}                       & \multicolumn{2}{c}{\textbf{Final Log PPL ($\downarrow$)}} & \multicolumn{2}{c}{\textbf{Final BLEU-1/2 ($\uparrow$)}} & \multicolumn{2}{c}{\textbf{Forgetting ($\downarrow$)}} & \multicolumn{2}{c}{\textbf{Final Log PPL ($\downarrow$)}} & \multicolumn{2}{c}{\textbf{Final BLEU-1/2 ($\uparrow$)}} & \multicolumn{2}{c}{\textbf{Forgetting ($\downarrow$)}} \\ \cmidrule(r){1-2} \cmidrule(lr){3-4} \cmidrule(lr){5-6} \cmidrule(lr){7-8} \cmidrule(lr){9-10} \cmidrule(lr){11-12} \cmidrule(lr){13-14}
& \textbf{Memory sizes}                  & \textbf{1,000}        & \textbf{10,000}       & \textbf{1,000}       & \textbf{10,000}      & \textbf{1,000}     & \textbf{10,000}    & \textbf{1,000}    & \textbf{10,000}    & \textbf{1,000}       & \textbf{10,000}      & \textbf{1,000}     & \textbf{10,000}    \\ 
\midrule
\multirow{9}{*}{\rotatebox[origin=c]{90}{\textbf{COCO-shift}}} 

& \textbf{Vanilla}                       & \multicolumn{2}{c}{5.040}                     & \multicolumn{2}{c}{25.96 / 1.29}              & \multicolumn{2}{c}{0.540}      & \multicolumn{2}{c}{5.193}              & \multicolumn{2}{c}{25.19 / 1.81}              & \multicolumn{2}{c}{0.612}               \\
& \textbf{ER}                            & \textbf{3.316}               & \textbf{2.430}\footnote{We updated the result of VLBERT with ER on COCO-shift dataset since the previous version of the paper. The previously reported results are 3.152 and 2.307 for memory sizes 1,000 and 10,000 respectively due to inconsistent experiment setups.}                 & \textbf{45.80} / 20.50          & 58.06 / 32.22          & 0.026             & -0.142             & \textbf{3.154}             & 2.426              & \textbf{49.52} / \textbf{24.56}          & 61.01 / \textbf{35.45}          & -0.069             & \textbf{-0.154}             \\
& \textbf{AGEM}                          & 3.478                 & 3.269                 & 37.94 / 13.56          & 40.21 / 15.64          & 0.235              & 0.145              & 3.411             & 3.227              & 38.71 / 14.72          & 40.28 / 15.87          & 0.361              & 0.257              \\
& \textbf{ER+MIR}                        & 3.342                 & 2.469                 & \textbf{45.80} / \textbf{20.87}          & 58.00 / 32.33          & 0.012             & -0.133             & 3.162             & \textbf{2.374}              & 48.77 / 23.72          & 60.79 / 35.10          & \textbf{-0.076}             & -0.147             \\
& \textbf{ER+MIR$_{max}$} &        3.344               & 2.473                 &          45.53 / 20.23            & \textbf{58.14} / \textbf{32.48}          &        \textbf{-0.056}            & \textbf{-0.153}             &      3.218             &        2.378        &          48.03 / 23.10            &          \textbf{61.06} / 35.24            &      -0.040              &         -0.140           \\ 
& \textbf{ER-10k$_{text-only}$}        & \multicolumn{2}{c} {3.108}                     & \multicolumn{2}{c}{47.99 / 22.51}              & \multicolumn{2}{c}{-0.128}    & \multicolumn{2}{c} {3.106}              & \multicolumn{2}{c}{48.07 / 22.57}              & \multicolumn{2}{c}{-0.112}                \\ \cmidrule(r){2-14}
 & \textit{Non-continual Learning Comparators} & & & & & & & & & & & & \\
& \textbf{Offline (one pass)}                    & \multicolumn{2}{c}{1.610}                     & \multicolumn{2}{c}{65.27 / 39.61}             & \multicolumn{2}{c}{-}                   & \multicolumn{2}{c}{1.783}              & \multicolumn{2}{c}{61.74 / 36.03}             & \multicolumn{2}{c}{-}                   \\
& \textbf{Offline}                   & \multicolumn{2}{c}{1.443}                     & \multicolumn{2}{c}{68.93 / 44.16}             & \multicolumn{2}{c}{-}                   & \multicolumn{2}{c}{1.503}              & \multicolumn{2}{c}{67.53 / 42.79}             & \multicolumn{2}{c}{-}                   \\

\midrule
\addlinespace
\addlinespace
\midrule
\multirow{9}{*}{\rotatebox[origin=c]{90}{\textbf{Flickr-shift}}}
& \textbf{Vanilla}        & \multicolumn{2}{c}{5.691}                     & \multicolumn{2}{c}{25.01 / 3.01}              & \multicolumn{2}{c}{0.456}           & \multicolumn{2}{c}{6.107}              & \multicolumn{2}{c}{24.77 / 3.09}              & \multicolumn{2}{c}{0.619}                    \\
& \textbf{ER}             & 5.016                 & \textbf{3.492}                 & \textbf{29.56} / \textbf{7.96}           & 40.23 / 16.80          &        0.229            &       0.023             & 4.949             &         \textbf{3.197}           & 31.32 / 9.58           &      \textbf{44.34} / 20.73          &          0.237          &     0.021       \\
& \textbf{AGEM}           & \textbf{4.493}                 & 4.393                 & 28.43 / 6.52           & 28.97 / 7.35           &         \textbf{0.004}           &      \textbf{-0.056}              & 5.246             & 5.072              & 25.18 / 3.63           & 25.28 / 3.53           &         \textbf{0.108}           &   0.096                 \\
& \textbf{ER+MIR}         & 5.118                & 3.504                 & 29.40 / 7.48           & \textbf{40.27} / \textbf{16.81}          &          0.268          &   -0.020                 & 4.949             & 3.211              & 31.64 / 9.80           & 44.30 / \textbf{20.83}         &       0.188             &     0.001               \\
& \textbf{ER+MIR$_{max}$} & 5.057                 & 3.555                 & 29.43 / 7.51           & 40.04 / 16.64          &         0.233           &       0.009         &     \textbf{4.788}              &    3.226                &        \textbf{31.72} / \textbf{9.89}          &         43.51 / 19.95             &     0.191               &     \textbf{-0.015}               \\  
& \textbf{ER-10k$_{text-only}$}        & \multicolumn{2}{c}{3.958}     &           \multicolumn{2}{c} {35.34 / 12.06}              &\multicolumn{2}{c} {0.070}            & \multicolumn{2}{c} {3.461}              & \multicolumn{2}{c} {39.07 / 16.71}              & \multicolumn{2}{c} {-0.008}                    \\ \cmidrule(r){2-14}
 & \textit{Non-continual Learning Comparators} & & & & & & & & & & & & \\
 & \textbf{Offline (one pass)}     & \multicolumn{2}{c}{2.590}                     & \multicolumn{2}{c}{47.08 / 21.88}             & \multicolumn{2}{c}{-}                   & \multicolumn{2}{c}{2.640}              & \multicolumn{2}{c}{47.30 / 22.56}             & \multicolumn{2}{c}{-}                   \\
& \textbf{Offline}    & \multicolumn{2}{c}{2.025}                     & \multicolumn{2}{c}{57.13 / 32.29}             & \multicolumn{2}{c}{-}                   & \multicolumn{2}{c}{2.025}              & \multicolumn{2}{c}{57.10 / 32.25}             & \multicolumn{2}{c}{-}                   \\

\bottomrule
\bottomrule
\end{tabular}%
}
\caption{\textbf{
Comparison of various training algorithms across two base-models (VLBERT and LXMERT) on the regular test set of COCO-shift and Flickr-shift.}
Here, PPL is Perplexity, BLEU-1/2 denotes BLEU-1 and BLEU-2 respectively,
metrics with ($\uparrow$) imply higher is better, similarly, metrics with ($\downarrow$) imply lower is better.
Best performance for each metric in a dataset is emphasized.
}
\label{tab:coco_overall}
\end{table*}
\end{savenotes}

\begin{figure*}[t]
    \centering
    \subfloat[][Comparison across methods, Noun-Verb ]{\includegraphics[width=0.5\textwidth]{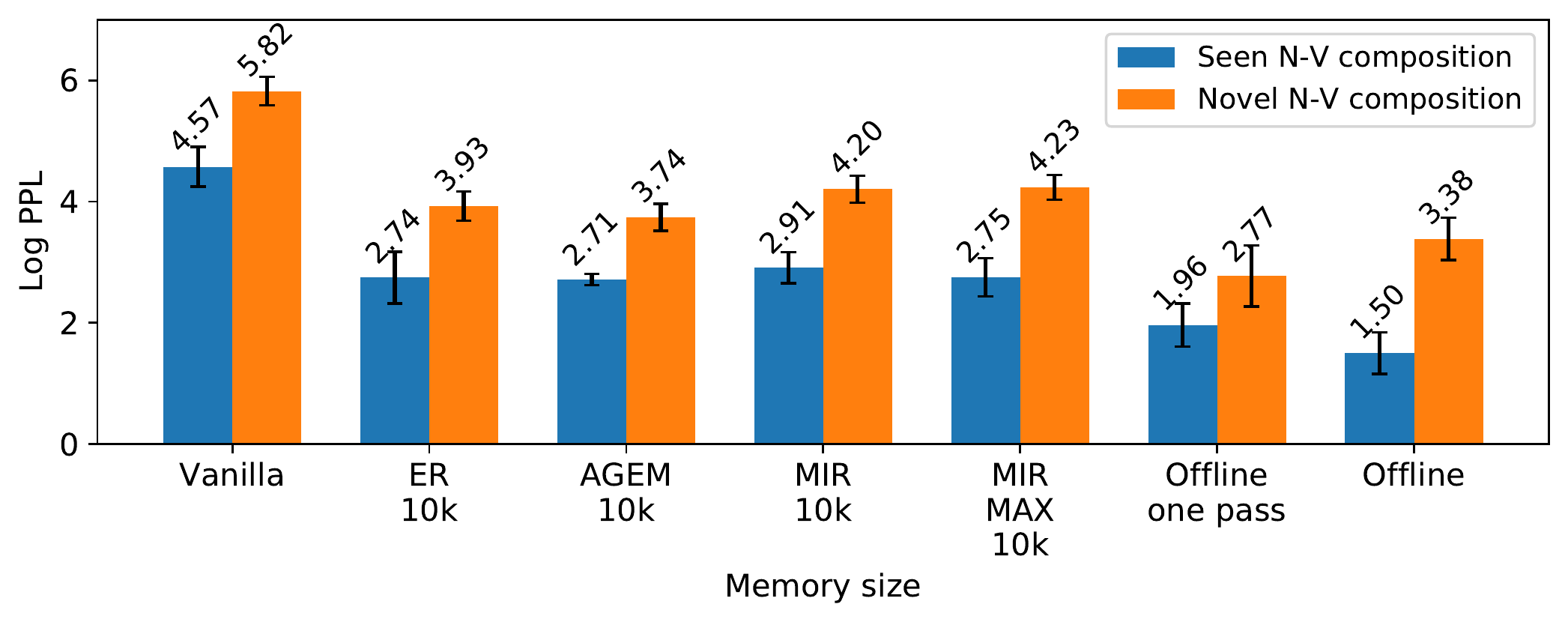}}
    \subfloat[][Comparison across methods, Noun-Adjective ]{\includegraphics[width=0.5\textwidth]{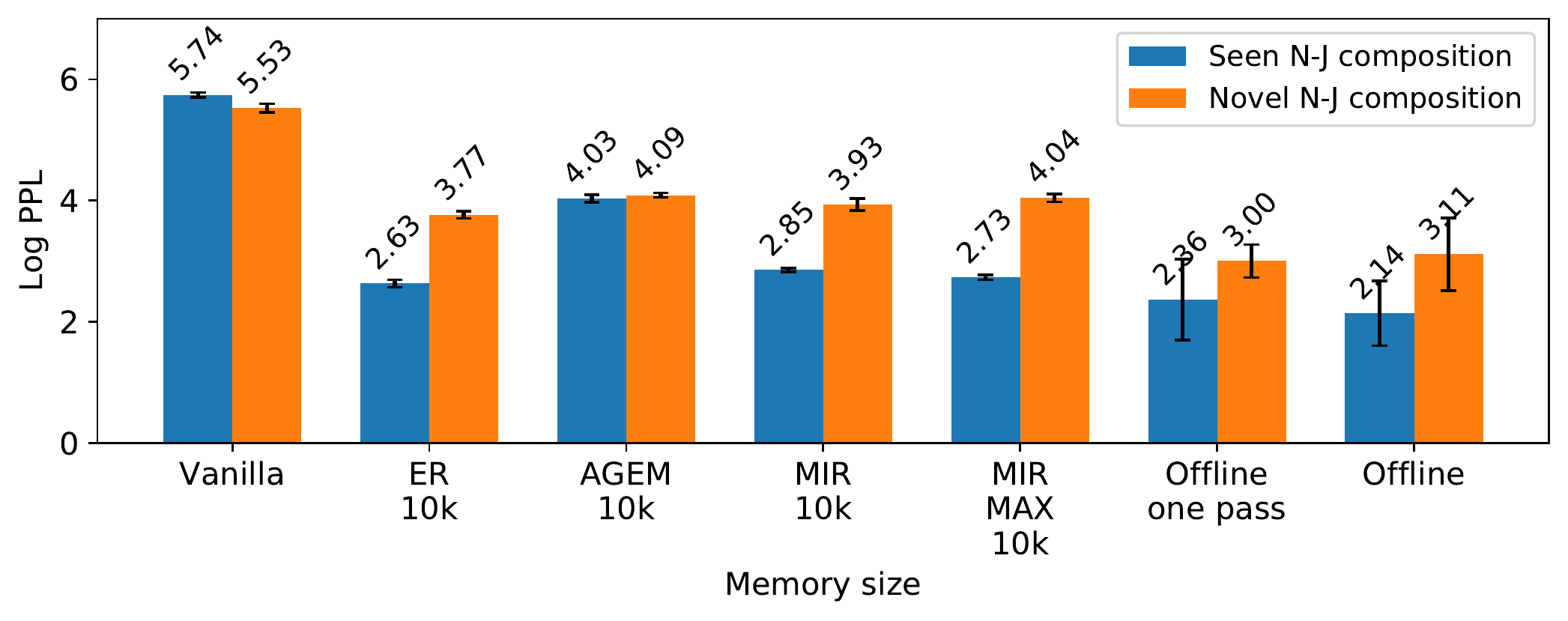}}

    \subfloat[][Comparison across memory sizes, Noun-Verb ]{\includegraphics[width=0.5\textwidth]{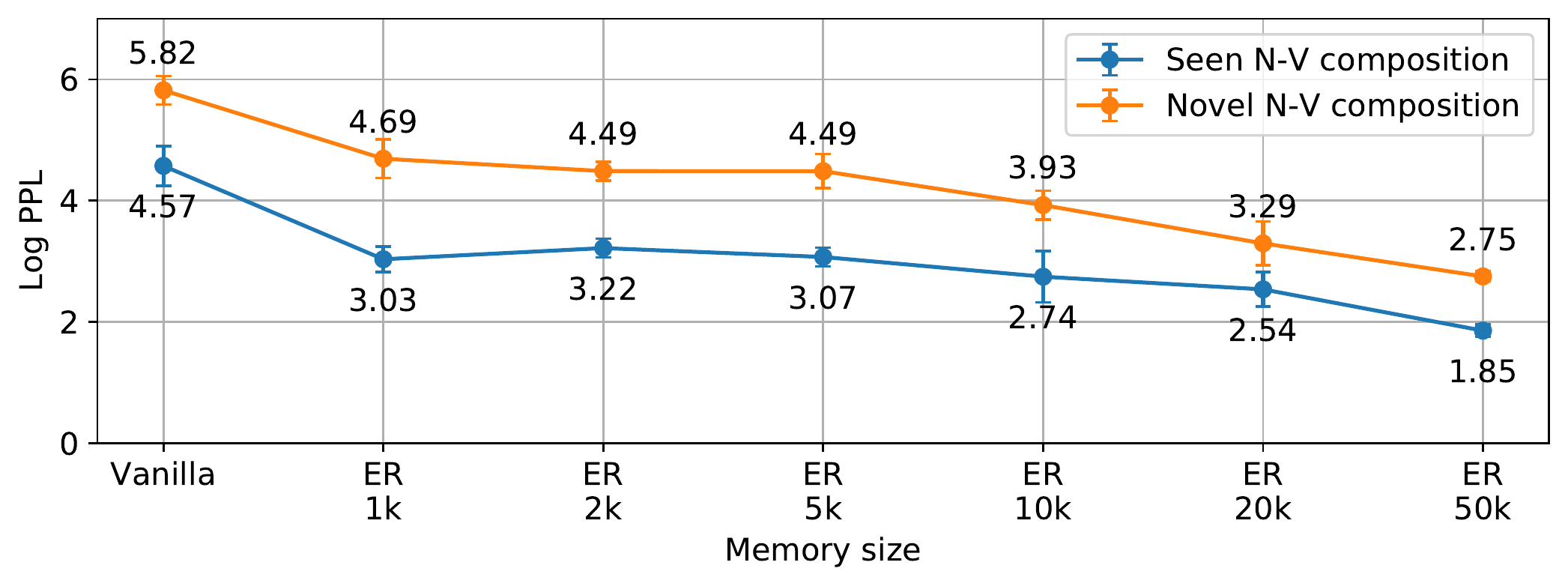}}
    \subfloat[][Comparison across memory sizes, Noun-Adjective ]{\includegraphics[width=0.5\textwidth]{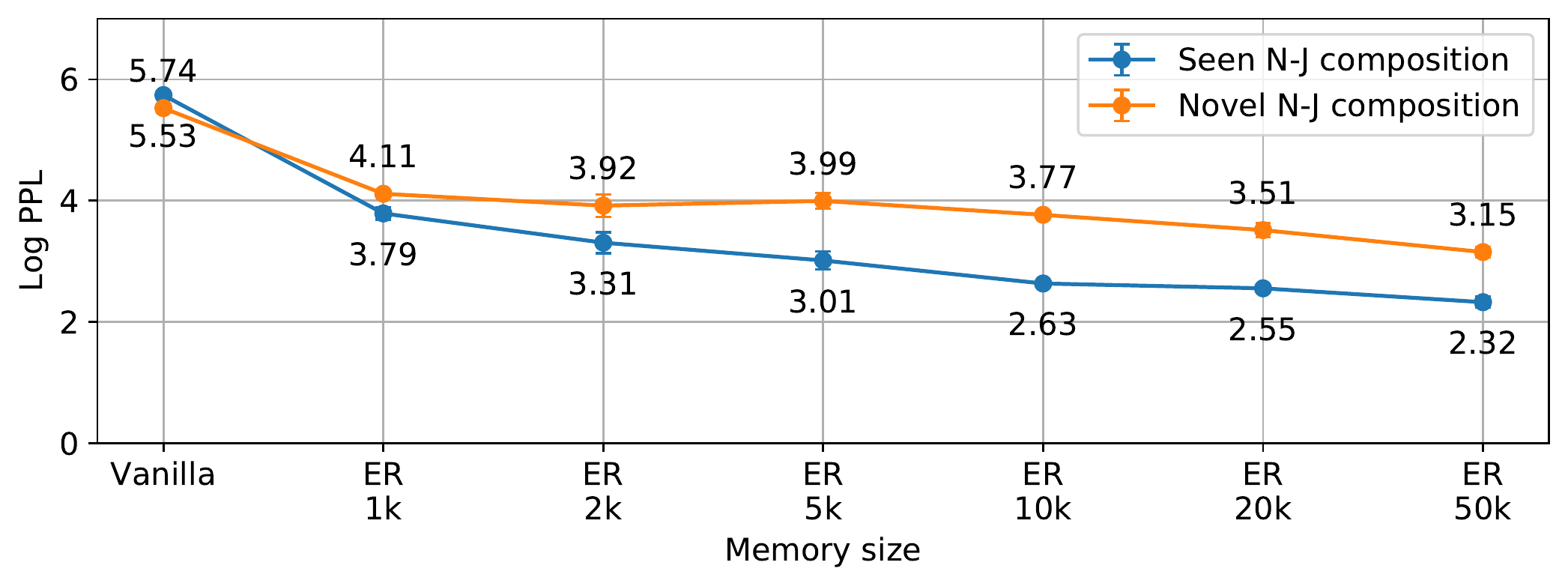}}
    \vspace{-0.1cm}
    \caption{
    \textbf{Results for Compositional Generalization.}
    We report perplexity of seen and novel compositions across methods (a),(b) and across memory sizes (c),(d) on COCO-shift dataset on noun-verb compositions and noun-adjective compositions separately.
    We first average the perplexity over examples for each composition individually, then compute the mean over these averaged scores over the set of compositions.
    }
    \label{fig:sn_bars}
\end{figure*}

\section{Experiments}
\label{sec:main_expt}
With the \vcoll{} task formulation in place, we study:
(i) performance of continual learning algorithms on \vcoll{}.
(ii) effect of the large search space on memory-based continual learning algorithms.
(iii) performance on generalizing to novel compositions. 
(iv) effect of replay memory management strategies.
We first describe our implementation details and metrics,
and present our results with findings.

\label{ss:esetup}
\noindent
\textbf{Implementation Details.}
For both VLBERT and LXMERT, 
we use a transformer with 6 layers, with 384 hidden units and 6 attention heads each. Note that all the parameters are learned from scratch without using pretrained weights.
For all continual learning algorithms, we use a memory size of $1k$ and $10k$, corresponding to nearly 
$0.2\%$ and $2\%$ of data for the two datasets. 
We report average over 3 runs with the fixed stream and the same set of random seeds.
See Appendix for more details.

\para{General Evaluation Metrics.}
We employ Perplexity (PPL) as our primary metrics (lower is better)
~\cite{mikolov2011empirical}.
Given a masked text span $W {=} [w_1,...,w_N]$ and the model's prediction probability output $P(W)$, the log-perplexity is defined as, $PPL(W)=- \frac{1}{N} \log P(W)$.
We report the perplexity in the log scale. 
Besides, we also use sentence-level BLEU~\cite{papineni2002bleu}.


\subsection{Study of Continual Learning}

To analyze the continual learning ability of our model, we use two metrics: 
(i) final perplexity and BLEU: the test set perplexity and BLEU scores
when the training ends
(ii) forgetting metric: the averaged perplexity increase over all tasks when the training ends compared to the checkpoints (almost) all of training data of a given task is observed. 
Mathematically, the forgetting metric is calculated as 
$f_{avg} = \frac{1}{|\mathcal{T}|} \sum_{k \in \mathcal{T}}  PPL_T(D_k) - PPL_{c_k}(D_k)$, where
$c_k = \argmin_{c_i \in \mathcal{C}} PPL_{c_i}(D_k)$.
$\mathcal{T}$ is the set of all tasks, and $\mathcal{C}$ is the set of all checkpoints;
$PPL_{c_k}(D_k)$ is the averaged perplexity over all test examples of task $k$ at the checkpoint $c_k$, and $T$ is the time step when the training ends. 
We expect a well-performing method to achieve low final perplexity, high BLEU scores, and low forgetting.




    

\label{ss:results}


\begin{figure*}[tb]

    \centering
    \subfloat[][Task \texttt{dog}]{\includegraphics[width=0.25\textwidth]{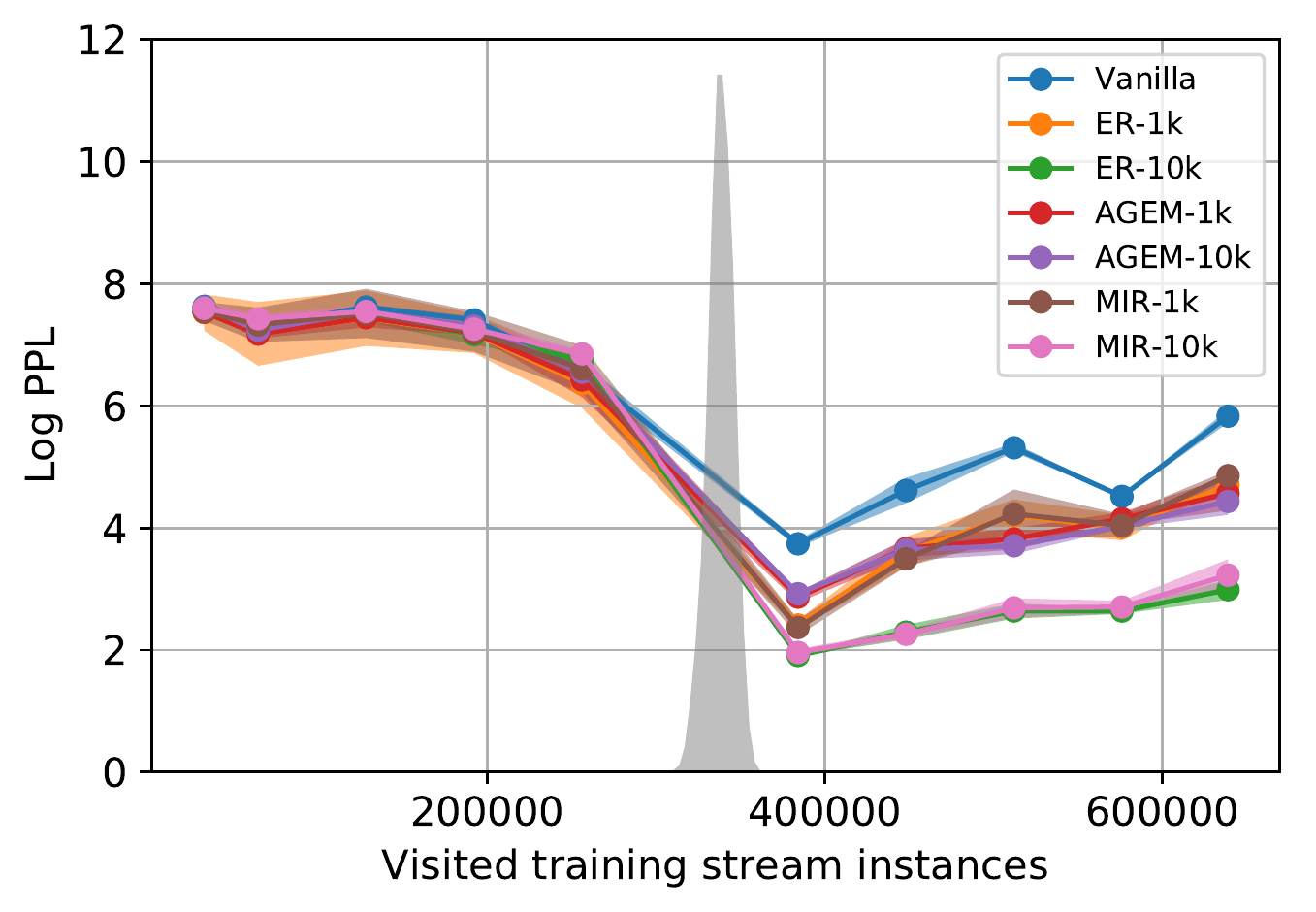}}
    \subfloat[][Task \texttt{sports ball}]{\includegraphics[width=0.25\textwidth]{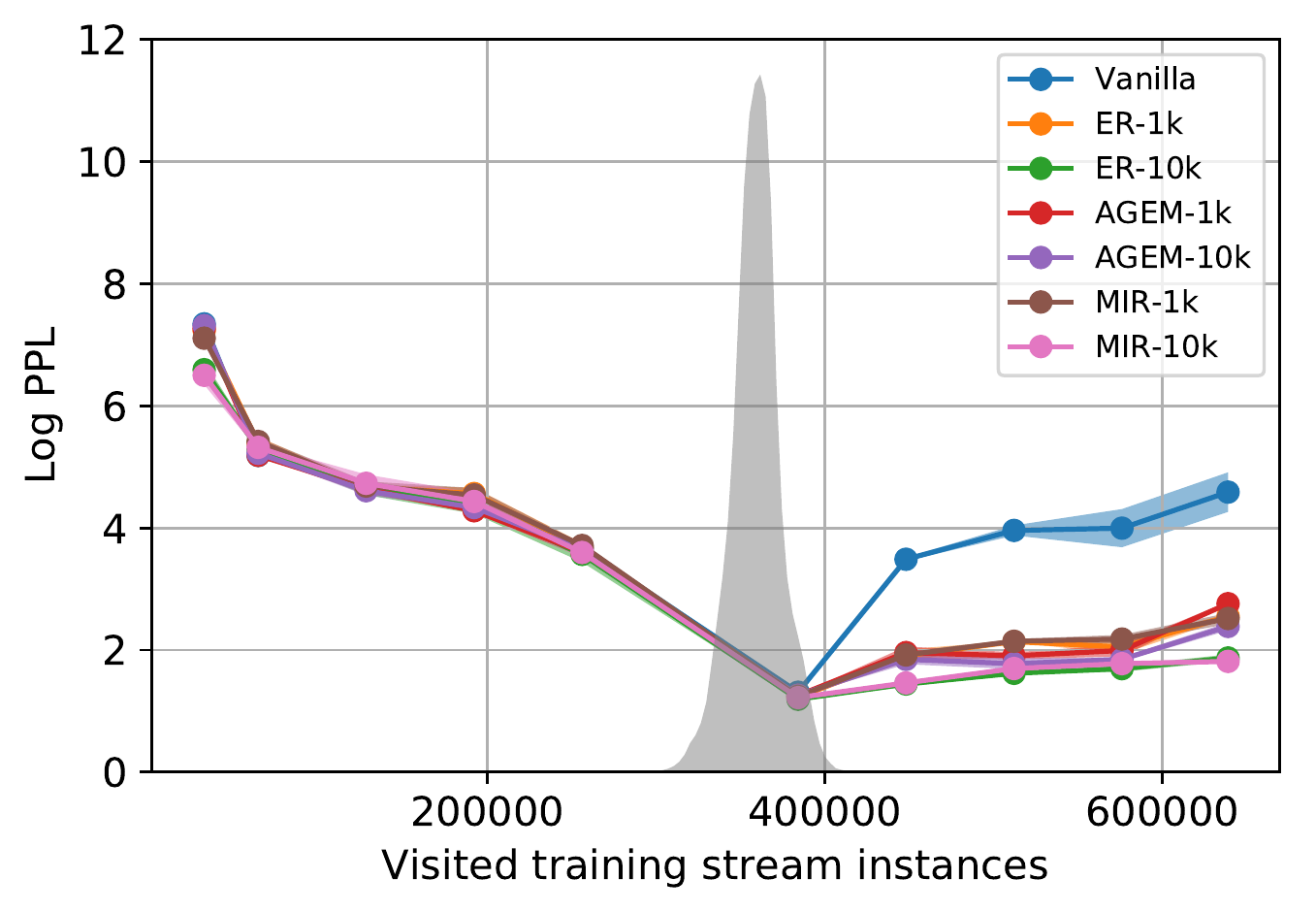}}
    \subfloat[][Task \texttt{apple}]{\includegraphics[width=0.25\textwidth]{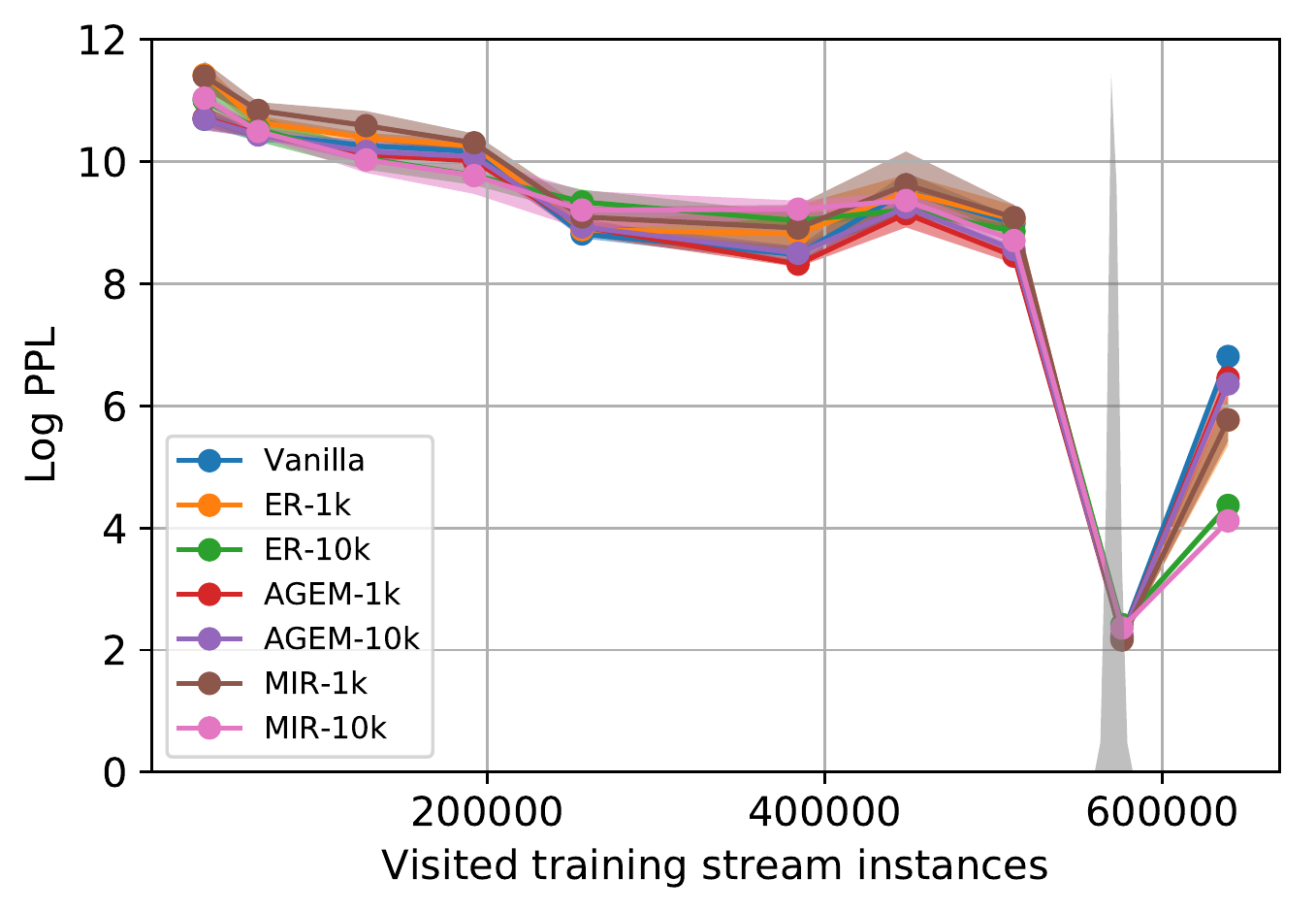}}
        \subfloat[][Task \texttt{book}]{\includegraphics[width=0.25\textwidth]{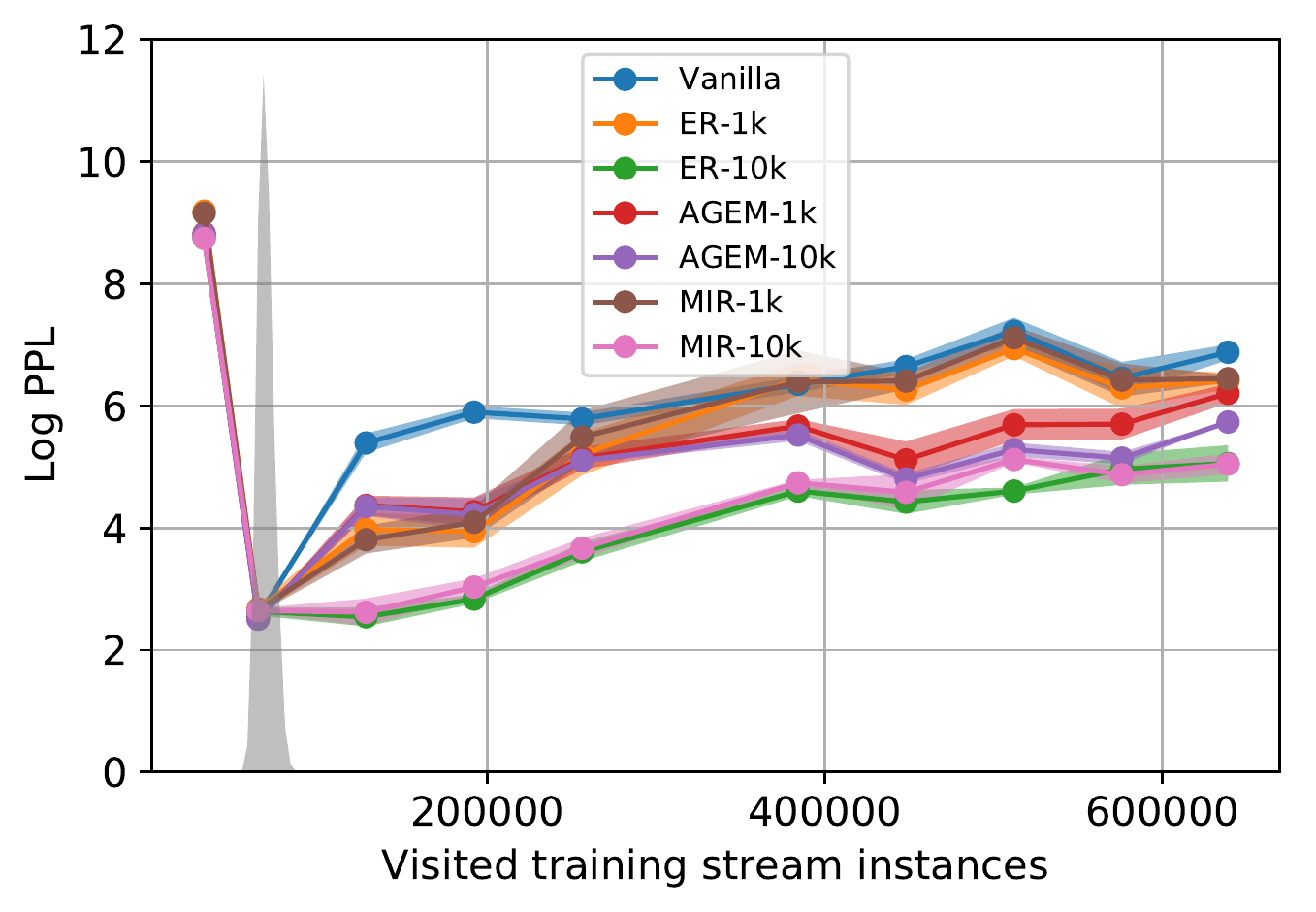}}

    \caption{
    \textbf{Comparing effects of continual learning algorithms}, exemplified with four tasks.
    $x$-axis is the training examples visited and $y$-axis is the perplexity.
    The gray-shaded regions in show the task distribution in the stream.
    }
    \label{fig:ppl_task_curves}
\end{figure*}



Tables \ref{tab:coco_overall} compares base-models with various continual strategies on the corresponding regular test splits of COCO-shift and Flickr-shift. We discuss our findings below.

\para{Non-stationary vs i.i.d Data Distribution.} 
Across both datasets, it is evident that models trained with the non-stationary distribution  (closer to what is observed in the real-world) largely under-perform compared to their i.i.d offline training counterpart at the single epoch setting ($20$-$40$ points difference in BLEU scores).
This emphasizes that catastrophic forgetting is prevalent in our constructed non-stationary data stream.

\para{Performance of Continual Learning Methods.} 
Despite its simplicity, ER achieves extremely competitive results, scoring within $1$-$2$ PPL of the best performing model.
While AGEM achieves very appealing final perplexity on Flickr-shift dataset at the 1$k$ memory setting (almost $0.5$ points better than alternatives), we find the corresponding BLEU is worse.
Given that perplexity evaluates over output probabilities, it is likely that AGEM makes less confident wrong predictions. 

We also find ER-MIR and its variant ER-MIR$_{max}$ occasionally outperforms ER, but the improvements are inconsistent over base-models and datasets. 
This is in stark contrast to continual learning benchmarks on image classification where algorithms like AGEM and ER-MIR achieve SoTA performance. In Fig.~\ref{fig:ppl_task_curves}(a)(b), we illustrate the change of perplexity over time for selected time steps in different methods.
We notice that with a memory size of $10k$, on average the forgetting metric for ER is close to or less than zero most of the time. This implies the performance of each task remains constant or improves over what was initially learned.

\para{Replay Memory Requirements Compared to Existing Benchmarks.}
It should be noted that even with a memory of $10k$ examples, the performance of continual learning algorithms are far from the i.i.d setting. 
In contrast to the 
popular continual learning benchmarks~\cite{kirkpatrick2017overcoming, zenke2017continual}, where storing only a few examples for each class is believed to be sufficient for a good performance~\cite{Chaudhry2019OnTE}. 

\para{Effect of Multi-Modal Data.} 
To decouple the gains obtained from visual and textual modality, we construct a text-only baseline by zeroing out the image-inputs in our base models and train using ER with memory size $10k$.
From Table ~\ref{tab:coco_overall}, we find across all cases, text-only baseline is outperformed by its multi-modal counterpart ($5$ points on BLEU) suggesting information from both image and captions is necessary to perform well on \vcoll.

Our findings underscore the challenges imposed by \vcoll{} and encourages closer examination towards existing continual learning benchmarks.



\subsection{Study of Compositional Generalization}
\begin{figure*}[tb]
 \vspace{-0.1cm}
    \centering
    \subfloat[][\texttt{sitting} in \texttt{sitting dog}]{\includegraphics[width=0.25\textwidth]{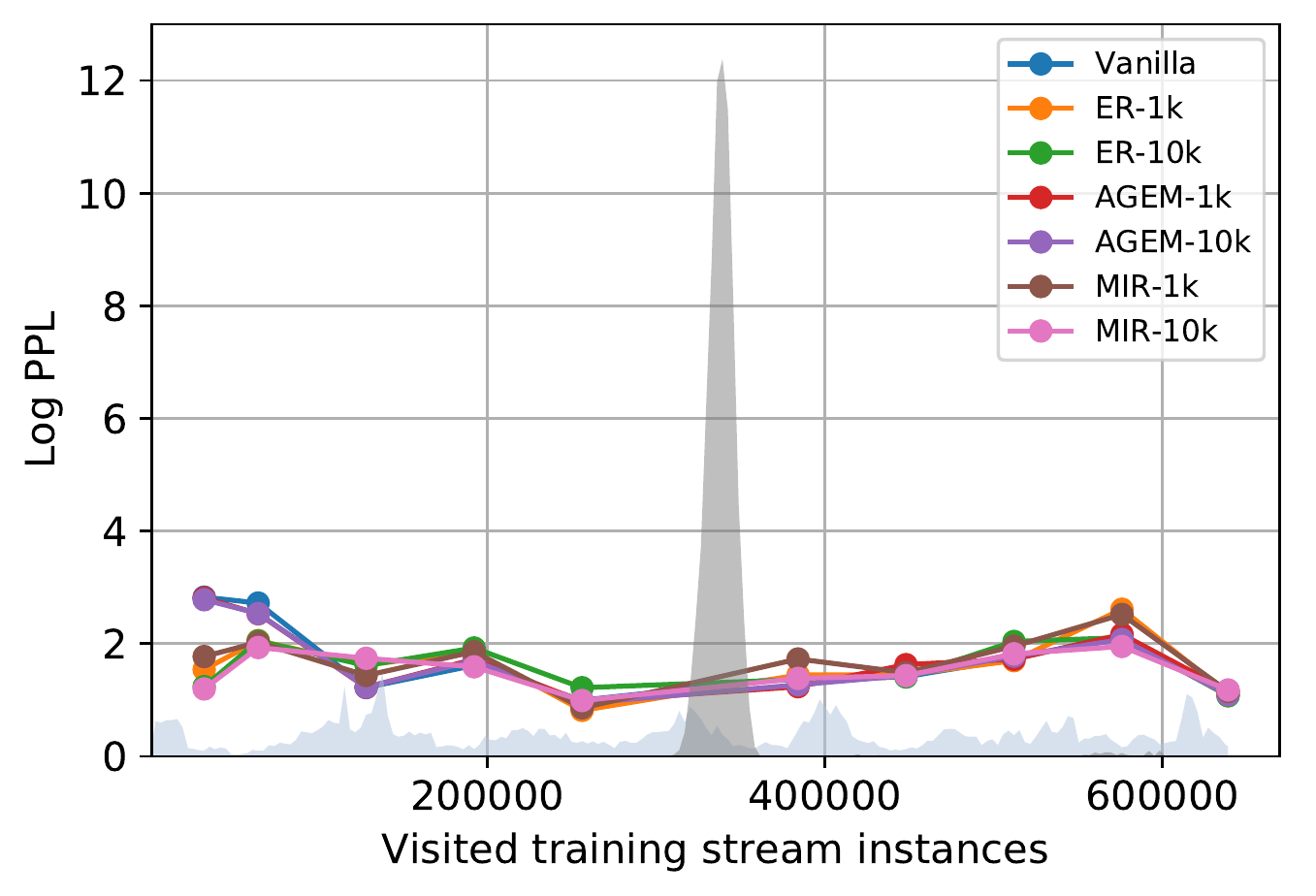}}
    \subfloat[][\texttt{running} in \texttt{running dog}]{\includegraphics[width=0.25\textwidth]{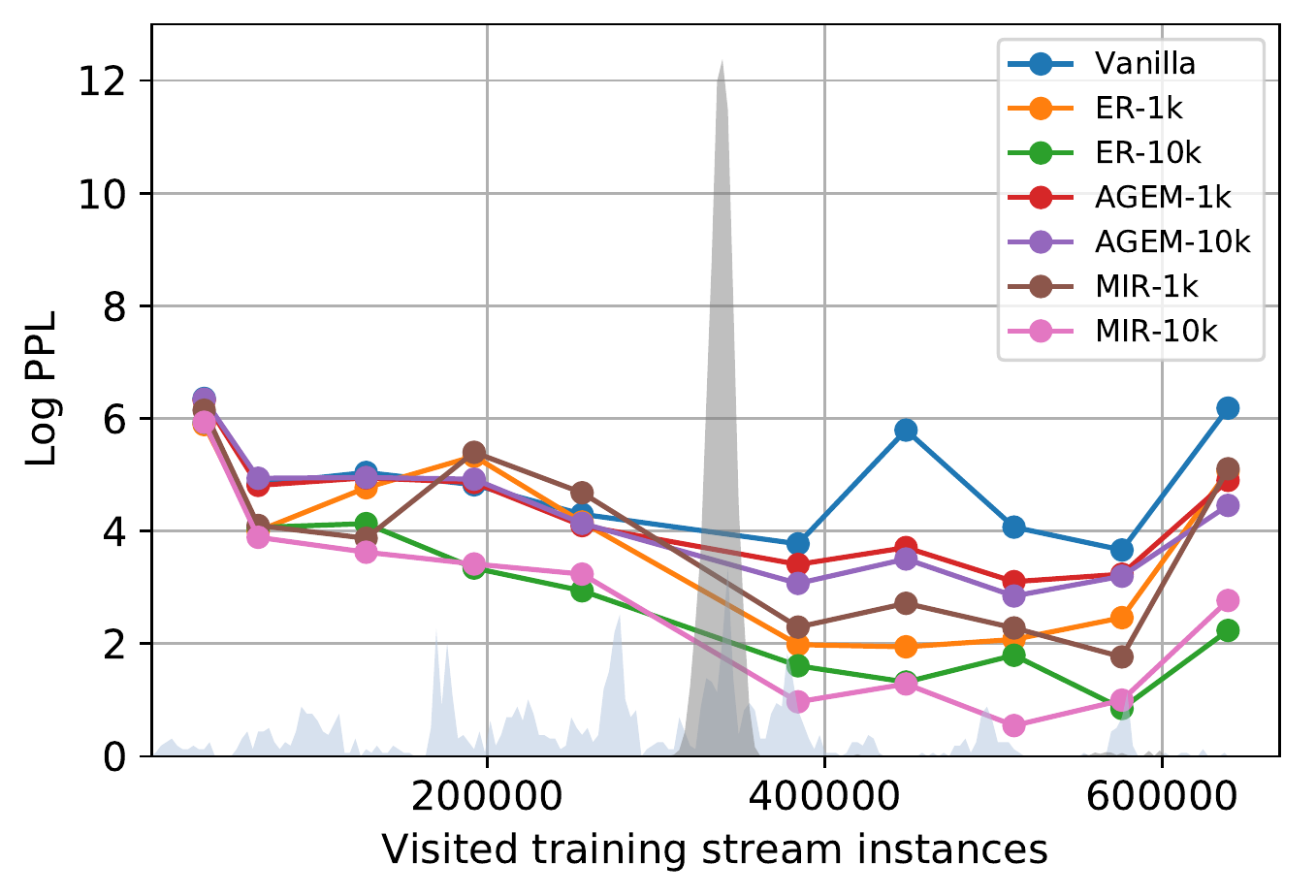}}
    \subfloat[][\texttt{black} in \texttt{black dog}]{\includegraphics[width=0.25\textwidth]{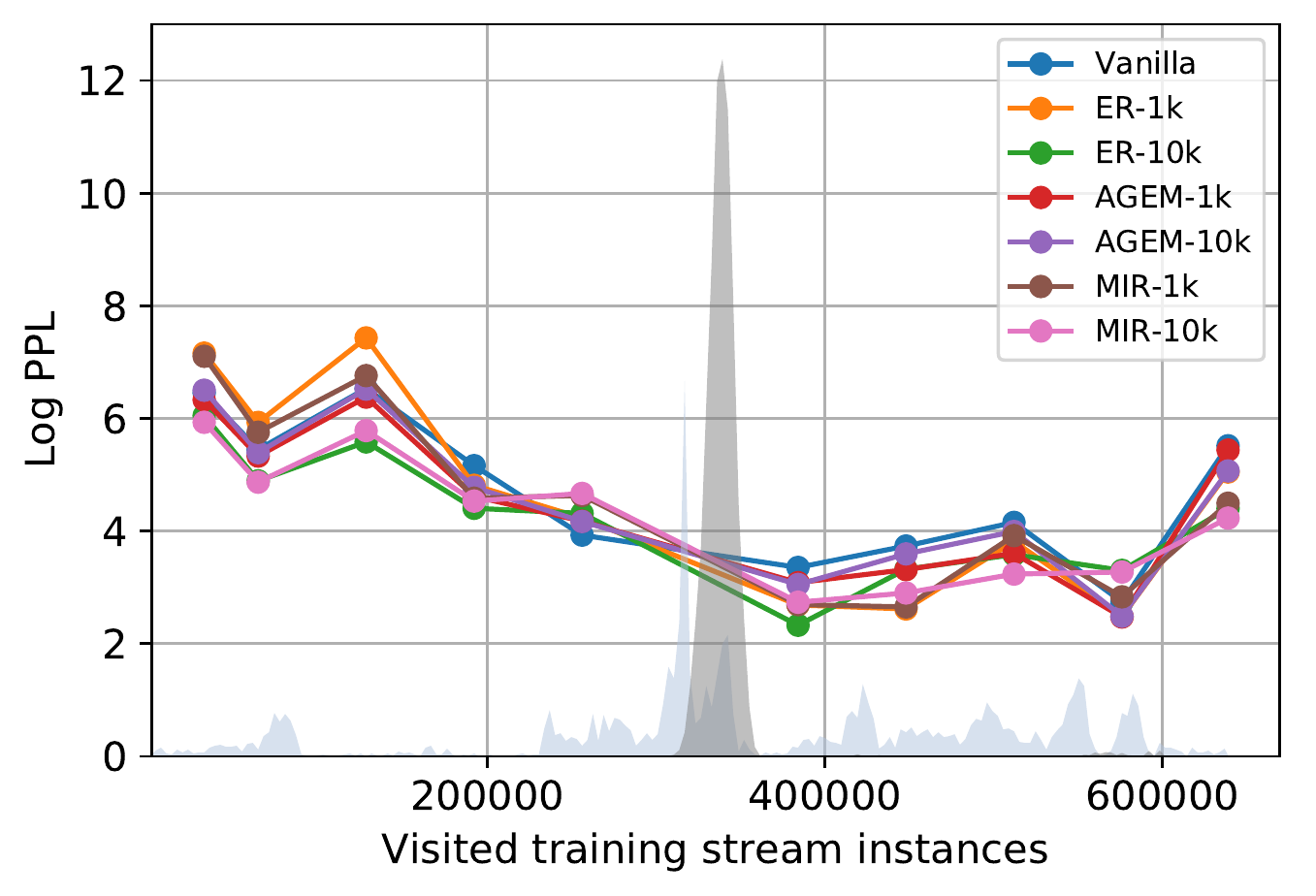}}
        \subfloat[][\texttt{big} in \texttt{big dog}]{\includegraphics[width=0.25\textwidth]{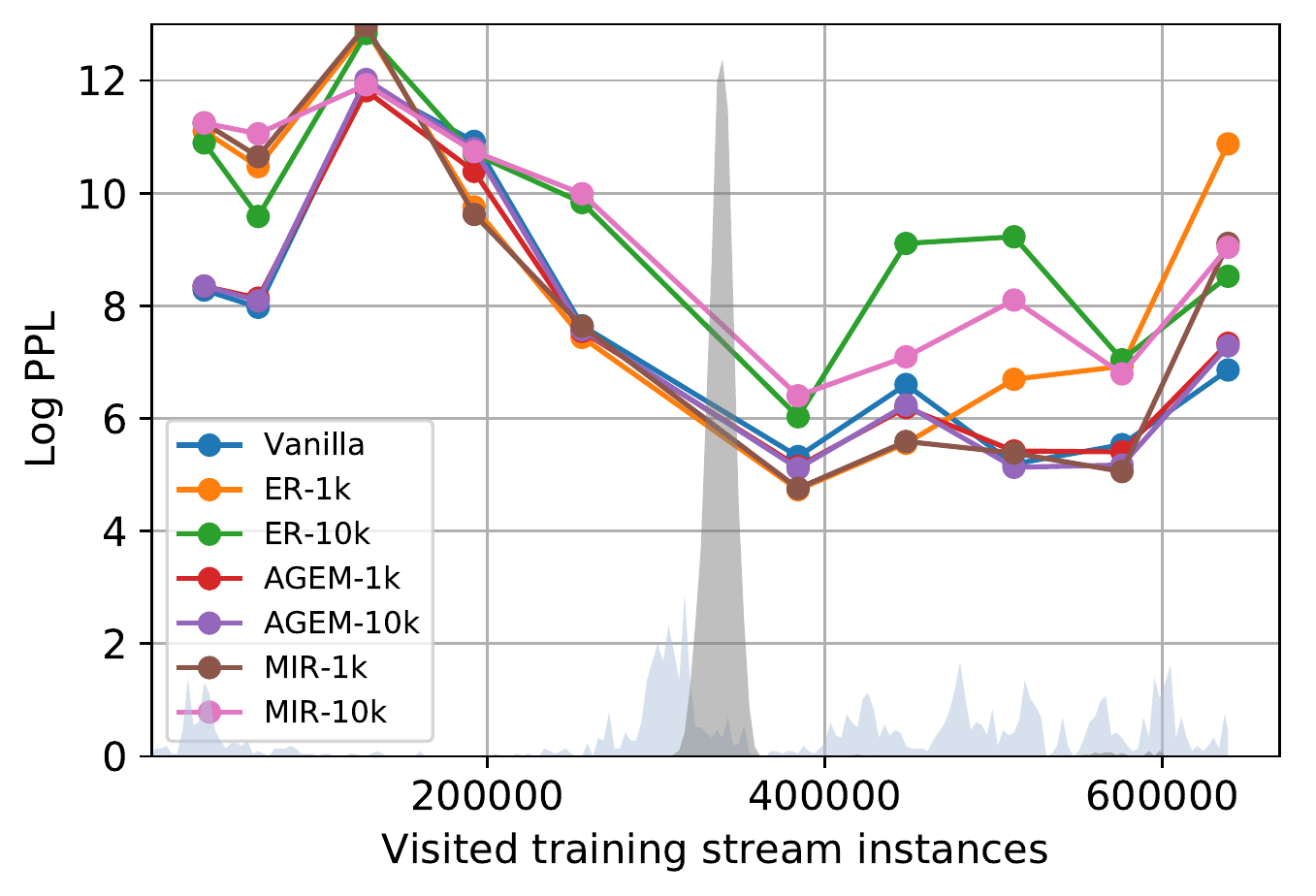}}
    \vspace{-0.1cm}
    \caption{
    \textbf{Analyzing forgetting in old noun-verb or noun-adjective compositions sharing the same noun}. 
    $x$-axis is the training examples visited and $y$-axis is the perplexity of the verb / adjective.
    The sharp gray-shaded regions are for the noun, while the light-blue regions near $x$-axis are for the adjectives.
    }
    \label{fig:ppl_task_curves}
\end{figure*}

To measure compositionality captured by models, in addition to a regular test set, we evaluate on the compositional test set which consists of novel noun-adjective and noun-verb pairs. 
We compare the performance with seen compositions sharing the same set of atomic words in the regular test set.
For a fair comparison with novel splits, we compare the performance on held-out novel pairs with a subset of the regular test-set sharing the same set of atomic words.

\para{Overall Compositional Generalization Results}. We plot the results in Figure~\ref{fig:sn_bars}. We note that the performance on novel compositions drops across all cases implying composition generalization is  very difficult for visual language transformers.
Interestingly, offline (one pass) outperforms offline training on novel compositions, suggesting the latter is over-fitting to the ``seen'' case.


\para{Analyzing Performance on Old Compositions.}
In an online setting, 
we further probe the model's
predictive performance on old seen compositions. 
Interestingly, we find that the performance is largely dependent on the atomic words used in the compositions. 
For instance, the performance drop on predicting ``black'' in the composition ``black dog'' is relative small (Figure \ref{fig:ppl_task_curves}(c)) compared to predicting ``big'' in ``big dog'' (Figure \ref{fig:ppl_task_curves}(d)).

\begin{figure}
 \vspace{-0.3cm}
    \centering
    \subfloat[][MSCOCO, VLBERT]{\includegraphics[width=0.23\textwidth]{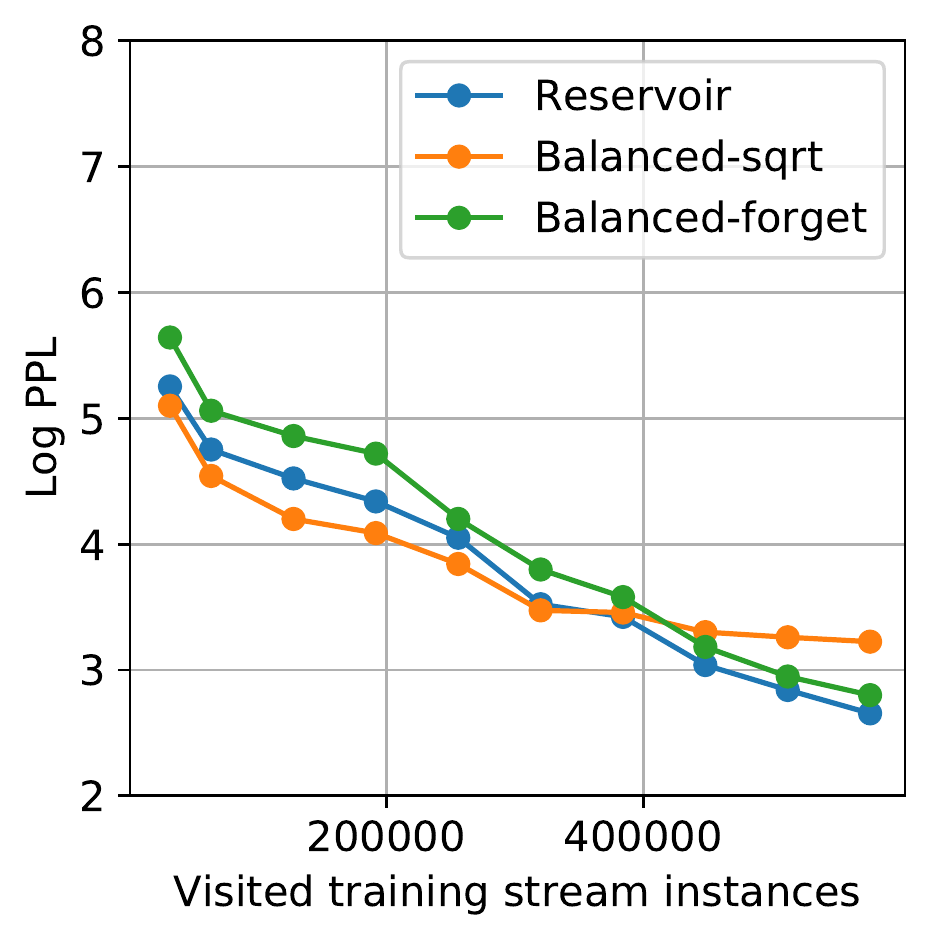}}
    \subfloat[][Flickr, LXMERT]{\includegraphics[width=0.23\textwidth]{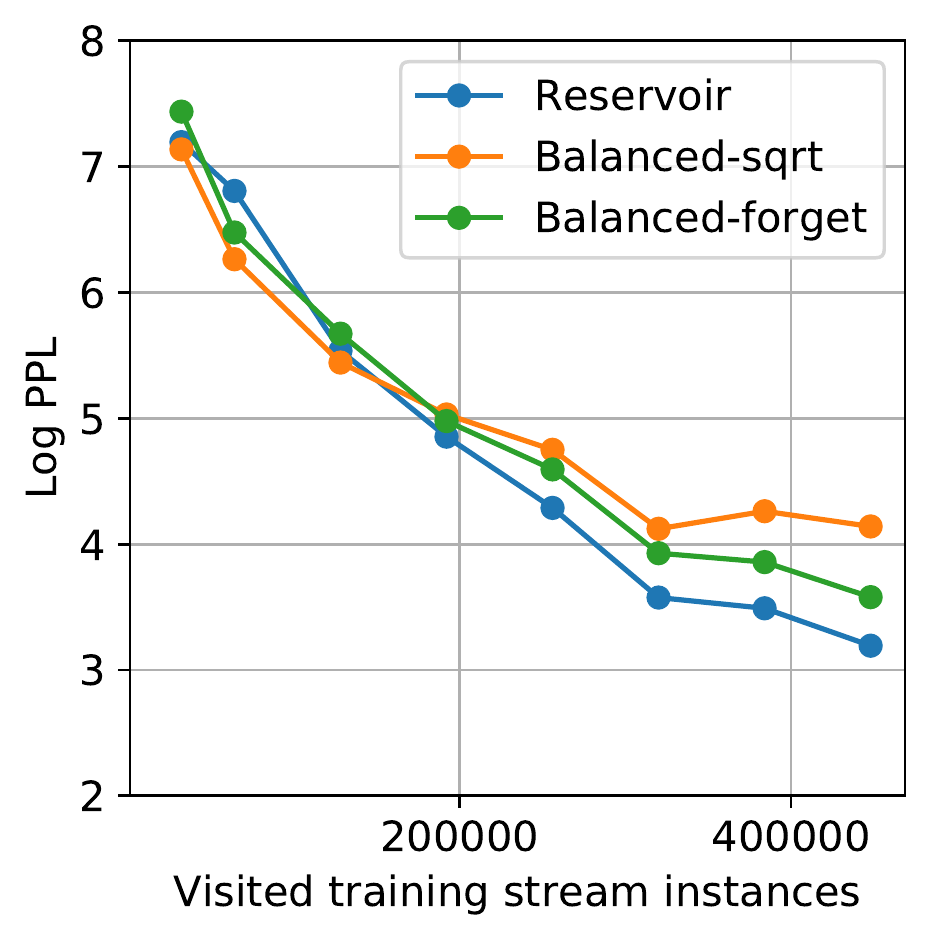}}
    \vspace{-0.2cm}
    \caption{\textbf{Effect of memory management strategies, studied with ER and a replay memory of $10k$ examples.} $x$-axis is the training step; $y$-axis is the perplexity.}
    \label{fig:attempts}
\end{figure}

\subsection{Study of Memory Management Strategies}
\label{ss:rep_mem_mgm}
We further study two memory scheduling strategies to account for a limited memory but large search space. Recall that the reservoir sampling applied our main experiments keeps each visited example has the same probability to be stored in the memory. 
We study two methods targeting storing more useful examples, aiming at:
(i) storing diverse compositions, and (ii) prioritizing likely forgotten words. 


We first propose target word distributions $p_{tgt}$ in the memory.
For (i), the target probability of each word is set proportional to the square root of its frequency in the visited stream.
Thus, highly frequent words would take a smaller portion 
compared to reservoir sampling where the word distribution in the memory is linear to its frequency in the visited stream,
leaving space for storing more diverse examples. 
We call this strategy \textbf{Balanced-sqrt}.
For (ii), the target probability is proportional to its frequency in the stream multiplied by its forgetting estimated during training ($\textit{i.e.,}$ loss increase). We call this strategy \textbf{Balanced-forget}.

For both strategies, given the word distribution in the memory $p_{mem}$ and target word distributions $p_{tgt}$, we 
minimize the KL-divergence $\mathrm{KL}(p_{mem}|| p_{tgt})$.
Thus, each time an example is received from the stream, 
we choose the memory example which if replaced causes the largest decrease in KL-divergence.
If there are no such memory examples that let KL-divergence decrease, we discard the example.



The results are compared in Figure~\ref{fig:attempts}.
We find that  
(i) diversifying storage improves performance at the early stage of the stream but not in the later stages; 
(ii) prioritizing words likely to be forgotten does not improve performance. 
Thus, future works should find a balance between storing more diverse or important examples and respecting original data distribution.

\section{Conclusion}
We propose \vcoll{}, 
a novel continual learning setup for visually grounded language acquisition.
\vcoll{} presents two main challenges: continual learning and compositionality generalization.
To facilitate study on \vcoll,  
we generate continuously shifting data-stream to construct two datasets, namely COCO-shift and Flickr-shift, and establish evaluation protocols. 
We benchmark our proposed datasets with extensive analysis using state-of-the-art continual learning methods. 
Experiments reveal that continual learning algorithms  struggle at composing phrases which have a very large search space, and show very limited generalization to novel compositions. Future works include looking into models and continual learning algorithms to better address the challenge.



\section*{Acknowledgements}
This research is based upon work supported in part by the Defense Advanced Research Projects Agency (DARPA) under Agreement No. HR00111990059, and the DARPA MCS program under Contract No. N660011924033 with the United States Office Of Naval Research. The views and conclusions contained herein are those of the authors and should not be interpreted as necessarily representing the official policies, either expressed or implied, of ODNI, IARPA, or the U.S. Government. We would like to thank all the collaborators in USC INK research lab for their constructive feedback on the work.

\bibliography{main}
\bibliographystyle{acl_natbib}

\newpage
\clearpage
\appendix

\begin{table*}[ht]
\centering

\scalebox{0.80}{\begin{tabular}{@{}lcccccccc@{}}
\toprule
\textbf{Dataset}        & \multicolumn{4}{c}{\textbf{COCO-shift}}                                   & \multicolumn{4}{c}{\textbf{Flickr-shift}}                                 \\ \midrule
\textbf{Method/Model}   & \multicolumn{2}{c}{\textbf{VLBERT}} & \multicolumn{2}{c}{\textbf{LXMERT}} & \multicolumn{2}{c}{\textbf{VLBERT}} & \multicolumn{2}{c}{\textbf{LXMERT}} \\ \midrule
\textbf{Memory sizes}   & \textbf{1,000}   & \textbf{10,000}  & \textbf{1,000}   & \textbf{10,000}  & \textbf{1,000}   & \textbf{10,000}  & \textbf{1,000}   & \textbf{10,000}  \\ \midrule
\textbf{iid-online}     & \multicolumn{2}{c}{0 h 59 min}      & \multicolumn{2}{c}{2 h 6 min}       & \multicolumn{2}{c}{0 h 35 min}      & \multicolumn{2}{c}{1 h 10 min}      \\
\textbf{Vanilla}        & \multicolumn{2}{c}{0 h 59 min}      & \multicolumn{2}{c}{2 h 0 min}       & \multicolumn{2}{c}{0 h 34 min}      & \multicolumn{2}{c}{1 h 12 min}      \\
\textbf{ER}             & 1 h 37 min       & 1 h 40 min       & 3 h 26 min       & 3 h 35 min       & 1 h 5 min        & 1 h 8 min        & 2 h 6 min        & 2 h 10 min       \\
\textbf{AGEM}           & 2 h 57 min       & 2 h 36 min       & 5 h 16 min       & 5 h 20 min       & 1 h 32 min       & 1 h 38 min       & 3 h 30 min       & 3 h 29 min       \\
\textbf{ER-MIR}         & 3 h 49 min       & 3 h 31 min       & 7 h 30 min       & 8 h 22 min       & 2 h 9 min        & 2 h 42 min       & 5 h 1 min        & 5 h 14 min       \\
\textbf{ER-MIR$_{max}$} & 3 h 29 min       & 3 h 30 min       & 8 h 7 min        & 8 h 20 min       & 2 h 19 min       & 2 h 49 min       & 4 h 58 min       & 5 h 8 min        \\ \bottomrule
\end{tabular}}
\caption{Average training time over a single pass of the stream.}
\label{tab:runtime}
\end{table*}

\section{Details of Dataset Construction}
\begin{table}[]
\centering
\scalebox{0.9}{\begin{tabular}{@{}cccc@{}}
\toprule
black cat   & big bird    & red bus     & small plane \\ 
eat man     & lie woman   & white truck & small cat   \\
brown dog   & big plane   & ride woman  & fly bird    \\
white horse & big cat     & blue bus    & small table \\
hold child  & stand bird  & black bird  & small dog   \\
white boat  & stand child & big truck   & eat horse   \\ \bottomrule
\end{tabular}}
\caption{24 compositions used for the compositional test split of COCO-split dataset.}
\label{tab:heldout_phrase_list}
\end{table}

\para{Heldout Phrases}. We put the complete list of 24 noun-verb and noun-adjective compositions used as novel compositions in Table~\ref{tab:heldout_phrase_list}, which are provided in~\cite{Nikolaus2019CompositionalGI}.

\section{Hyperparameter Settings}
Since the complete stream should not be assumed known apriori in the online learning setting, 
following prior work ~\cite{chaudhry2018efficient},
we use a small portion (10\%) of the training data and the validation set to perform hyperparameter search.
We use AdamW optimizer~\cite{Loshchilov2019DecoupledWD} throughout the experiements.
We tune the learning rate based on the validation performance on the Vanilla method averaged over 3 runs. For a fair comparison in the online learning setup, we use the same learning rates for all methods. The learning rate is selected from $\{2e^{-4}, 1e^{-4}, 5e^{-5}\}$ and decided as $1e^{-4}$. We set the batch size to 32 throughout the experiments. For ER, ER-MIR and ER-MIR$_{max}$, at each training iteration, we replay the same number of examples from the memory as the examples received from the stream (\textit{i.e.}, training batch size), following the convention in recent works~\cite{Aljundi2019OnlineCL, Chaudhry2019OnTE}. 

AGEM, unlike ER, requires a larger set of memory examples to compute regularizations. We set the numbers to 80 and 64 respectively for COCO-shift and Flickr-shift. While larger numbers can be preferable, it introduces significant time and resource consumption overhead in our problem setup, which is much larger in scale compared to existing continual learning benchmarks.

Similarly, ER-MIR and ER-MIR$_{max}$ introduce a hyperparameter for the size of the ``candidate set'' to retrieve examples that are most likely to be forgotten. We set the hyperparameters as 80 and 64 for COCO-shift and Flickr-shift respectively.

\section{Effect of Data Order}
Data order has been known to affect performance in continual learning~\citep{Greco2019PsycholinguisticsMC}. To illustrate this point, we conduct a simple experiment where the task centroids are sorted in the ascending or descending order.  We show the log perplexity metrics in Table~\ref{tab:task_order}. The results show a significant log-perplexity gap, which verifies that data order may significantly affect performance. We leave more in-depth analysis as future works. Note that throughout our main experiments, the task order is fixed as a random order.

\begin{table}[]
\centering
\scalebox{0.9}{\begin{tabular}{@{}lccc@{}}
\toprule
\textbf{Task-order / Method} & Vanilla & ER-1k & ER-10k \\ \midrule
\textbf{Asc. Frequency} & 4.31 & 3.22 & 2.42 \\
\textbf{Desc. Frequency} & 5.18 & 3.58 & 2.53 \\
\textbf{Random (main results)} & 5.04 & 3.15 & 2.31 \\ \bottomrule
\end{tabular}}
\caption{Performance with different task orders in COCO-shift and the VLBERT model.}
\label{tab:task_order}
\end{table}

\section{Infrastructures and Statistics}

We use PyTorch 1.0.0~\cite{NEURIPS2019_9015} with CUDA toolkit version 10.1. We train our models with NVIDIA 2080 Ti GPUs. Our VLBERT models have 23,564,040 trainable parameters and LXMERT models have 58,614,794 trainable parameters. We report the average training time over a single pass of the stream in table~\ref{tab:runtime}.


\end{document}